\documentclass[twocolumn]{svjour3}
\smartqed  
\usepackage{booktabs}
\usepackage{graphicx}
\usepackage{amsmath}
\usepackage{amssymb}
\usepackage{tabularx}
\usepackage{multirow}
\usepackage{algorithm}
\usepackage{algorithmic}
\usepackage{stackengine}
\usepackage{xcolor}
\usepackage{adjustbox}
\usepackage{colortbl}
\usepackage{ragged2e}
\usepackage{rotating,soul}
\usepackage[caption=false,font=footnotesize]{subfig}
\usepackage{tipa}
\usepackage{ulem}
\usepackage{color}
\usepackage{booktabs}
\usepackage{float}
\usepackage{bm}
\usepackage{bbding}
\usepackage{verbatim}
\usepackage{makecell}
\usepackage{overpic}
\usepackage{setspace}

\usepackage[numbers,sort]{natbib}
\bibpunct[,]{(}{)}{;}{a}{}{,}

\usepackage{pifont}%
\newcommand{\cmark}{\ding{51}}%
\newcommand{\xmark}{\ding{55}}%

\definecolor{mygray}{gray}{.93}
\definecolor{mygray1}{gray}{.99}
\newcommand{\myPara}[1]{\vspace{7pt}\noindent$\bullet$~\textbf{#1} \quad}

\def\ie{\textit{i.e.}}
\def\eg{\textit{e.g.}}
\def\etc{\textit{etc}}

\definecolor{linkcolor}{RGB}{255,0,0}
\definecolor{urlcolor}{RGB}{255,105,180}
\definecolor{citecolor}{RGB}{0, 80, 200}
\usepackage{hyperref}
\hypersetup{colorlinks=true,linkcolor=linkcolor,urlcolor=urlcolor,citecolor=citecolor}

\newcommand{\supp}[1]{\textcolor{magenta}{#1}}

\journalname{International Journal of Computer Vision}

\begin{document}\sloppy

\title{One-Shot Object Affordance Detection in the Wild}

\titlerunning{One-Shot Object Affordance Detection in the Wild}        
	\author{
		Wei Zhai$^{\textbf{1}\textbf{*}}$\and
		Hongchen Luo$^{\textbf{1}\textbf{*}}$\and 
		Jing Zhang$^\textbf{2}$\and 
		Yang Cao$^\textbf{1}$\and \\
		Dacheng Tao$^{\textbf{3}}$
	}
	
	\authorrunning{Zhai et al.} %

	\institute{
		$^{\textbf{1}}$University of Science and Technology of China, Hefei, China \\
		$^{\textbf{2}}$The University of Sydney, Sydney, Australia \\
		$^{\textbf{3}}$JD Explore Academy, China \\
		$^{\textbf{*}}$Equal contribution. This work was done during Wei Zhai's internship at JD Explore Academy.\\
	}


\date{Received: date / Accepted: date}

\maketitle

\begin{abstract}

Affordance detection refers to identifying the potential action possibilities of objects in an image, which is a crucial ability for robot perception and manipulation. To empower robots with this ability in unseen scenarios, we first study the challenging one-shot affordance detection problem in this paper, \ie, given a support image that depicts the action purpose, all objects in a scene with the common affordance should be detected. To this end, we devise a \textbf{O}ne-\textbf{S}hot \textbf{A}ffordance \textbf{D}etection \textbf{Net}work (OSAD-Net) that firstly estimates the human action purpose and then transfers it to help detect the common affordance from all candidate images. Through collaboration learning, OSAD-Net can capture the common characteristics between objects having the same underlying affordance and learn a good adaptation capability for perceiving unseen affordances. Besides, we build a large-scale Purpose-driven Affordance Dataset v2 (PADv2) by collecting and labeling $30$k images from $39$ affordance and $103$ object categories. With complex scenes and rich annotations, our PADv2 dataset can be used as a test bed to benchmark affordance detection methods and may also facilitate downstream vision tasks, such as scene understanding, action recognition, and robot manipulation. Specifically, we conducted comprehensive experiments on PADv2 dataset by including $11$ advanced models from several related research fields. Experimental results demonstrate the superiority of our model over previous representative ones in terms of both objective metrics and visual quality. The benchmark suite is available at  \href{https://github.com/lhc1224/OSAD_Net}{https://github.com/lhc1224/OSAD\_Net}.

\begin{figure}[t]
	\centering
		\includegraphics[width=8.41cm]{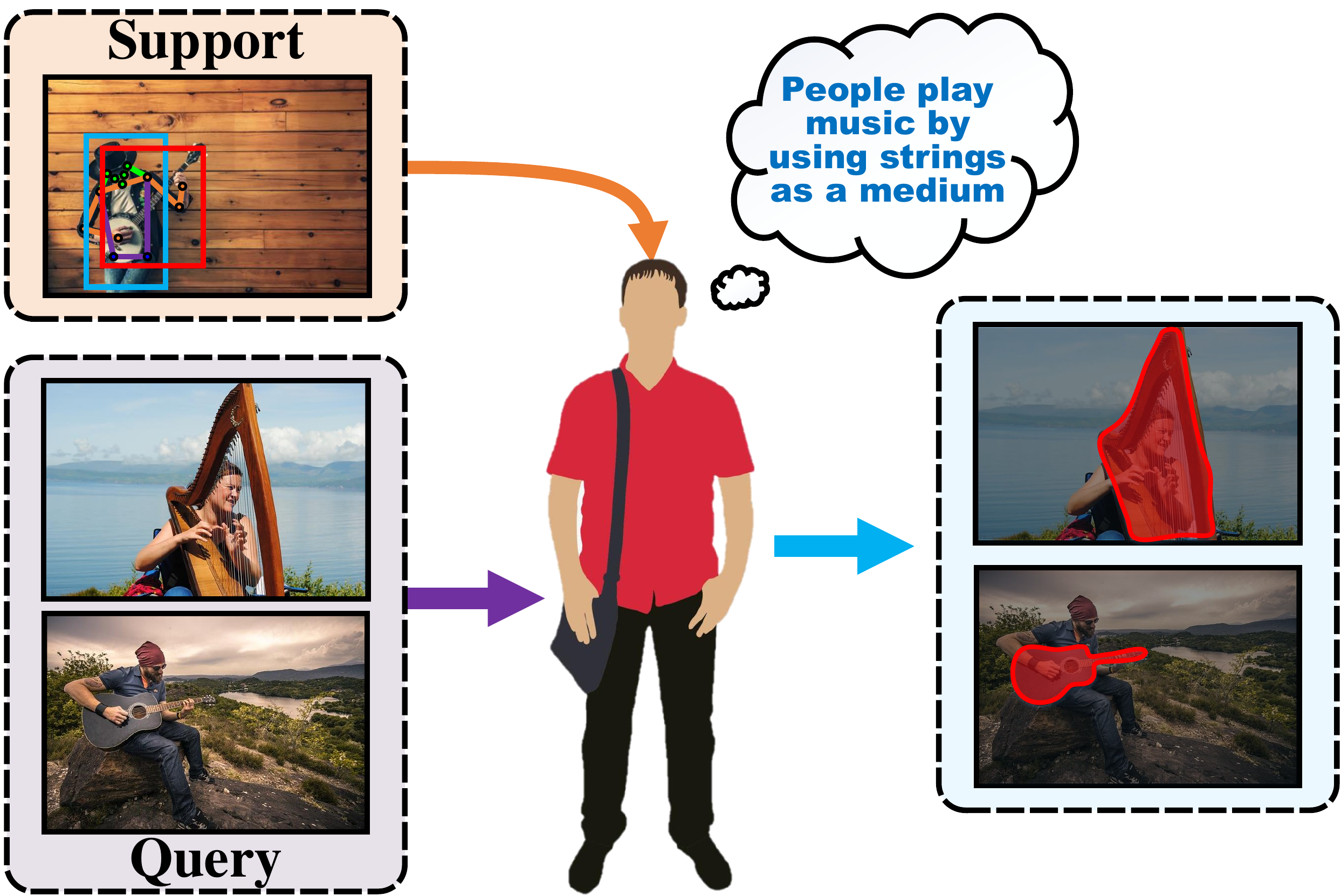}
	\caption{\textbf{Illustration of perceiving affordance.} Given a support image that depicts the action purpose, all objects in a scene with the common affordance could be detected.}
	\label{FIG:1}
\end{figure}

\begin{figure*}[t]
	\centering
		\includegraphics[width=17.41cm]{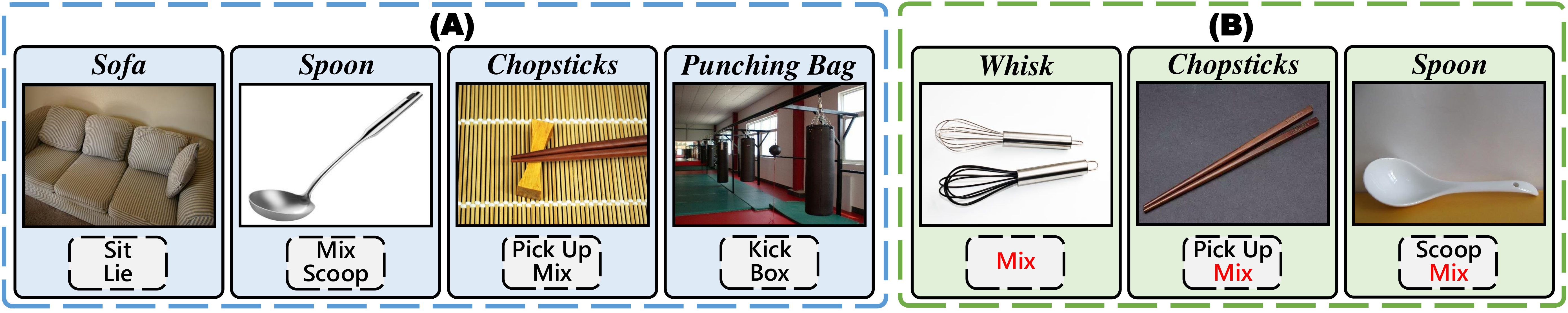}
	\caption{\textbf{Examples of multiple possibilities of object affordance.} Part (A) shows that objects usually have multiple affordances. Part (B) shows that the objects belonging to different semantic categories may have the same affordance.}
	\label{FIG:1_2}
\end{figure*}

\keywords{
Affordance Detectin \and One-Shot Learning \and Human Purpose Estimation and Transfer}
\end{abstract}

\section{Introduction}

The concept of affordance was proposed by the ecological psychologist Gibson \citep{gibson1977theory}. It describes how the inherent ``value'' and ``meanings'' of objects in an environment are directly perceived and explains how this information can be linked to the action possibilities offered to an organism by the environment \citep{hassanin2018visual}. In particular, perceiving the affordance of objects has a wide range of applications in a wide range of fields \citep{zhang2020empowering,DBLP:journals/corr/abs-1807-06775}, such as action recognition \citep{DBLP:journals/cviu/KjellstromRK11,qi2017predicting}, robot grasping \citep{yamanobe2017brief}, autonomous driving, and scene understanding  \citep{liu2019auto,vu2014predicting}.

Compared to the semantics of the object itself, affordance is an uncertain and dynamic property (\eg, a cell phone can both make phone calls and take pictures) that is closely related to the environment context and the possible interactions between the object and actors \citep{fang2020learning,hassan2015attribute,chuang2018learning,Zhu_2015_CVPR,wang2017binge}. Therefore, approaches that rely only on the construction of mapping relationships between the object structure and affordance labels of images in a fixed dataset do not have strong generalization capabilities, resulting in inferior affordance detection performance when the environment context or actor interactions change. Thus, it is expected that the model has the ability to adapt to unseen scenarios using a few samples or only one sample. Moreover, the ability of the model to rapidly adapt to changing scenes and generalize to unseen objects is of great significance in practical applications \citep{nagarajan2020learning,demo2vec2018cvpr,interaction-hotspots,chen2021deep}. 

To learn such a capability of perceiving affordance, we consider the challenging \textbf{one-shot affordance detection $\footnote[1]{``Detection'' refers to the pixel-wise detection task, which has also been used in the area of salient object detection.}$} task in this paper, \ie, given a support image that depicts the human action purpose, all objects in a scene with the common affordance should be detected (as shown in Fig. \ref{FIG:1}). Unlike the object detection/segmentation problem \citep{shaban2017one}, affordance and semantic categories of objects are highly inter-correlated but do not imply each other. An object may have multiple affordances (as shown in Fig. \ref{FIG:1_2} (A)), \eg, the sofa can be used to sit or lie down. An affordance category may cover more than one object category (as shown in Fig. \ref{FIG:1_2} (B)), \ie, whisk, chopsticks, and spoon all belong to the affordance category of ``mix''. The possible affordance depends on the human action purpose in real-world application scenarios. Directly learning the affordance from a single image without the guidance of action purpose makes the model focus on the statistically dominant affordances while ignoring other visual affordances that also coincide with the same action purpose.

To address this problem: 1) We try to find clear hints about the action purpose (\ie, via the subject and object locations and human poses \citep{chen2020recursive,wei2017inferring}) from a single support image, which implicitly defines the object affordance and thus can be used to reduce the affordance ambiguity of objects in candidate images. 2) We adopt collaboration learning to capture the inherent relationship between different objects to counteract the interference caused by visual appearance differences and improve generalization. Specifically, we devise a novel \textbf{O}ne-\textbf{S}hot \textbf{A}ffordance \textbf{D}etection \textbf{Net}work (\textbf{OSAD-Net}) to solve the problem. We take an image as support and a set of images ($5$ images in this paper) as a query, and the network first captures the human-object interactions from the support image via an action purpose learning (APL) module to encode the action purpose. Then, a mixture purpose transfer (MPT) module is devised to use the encoding of the action purpose to activate the features in query images that have common affordance. Finally, a densely collaborative enhancement (DCE) module is introduced to capture the intrinsic relationships between objects with the same affordance and suppress backgrounds irrelevant to the action purpose. In this way, our OSAD-Net can learn a good adaptation capability for perceiving unseen affordances.

Moreover, there is a gap between existing datasets and real-world application scenarios due to the limitation of their data diversity. The affordance detection model for scene understanding and general applications should be able to learn from the human-object interaction when the robot arrives at a new environment and search suitable objects as tools to complete specific tasks in the environment, rather than just finding objects with the same categories or similar appearance. To fill this gap, we propose the \textbf{P}urpose-driven \textbf{A}ffordance \textbf{D}ataset \textbf{v2} (\textbf{PADv2}), which contains $30$k diverse images covering $39$ affordance categories as well as $103$ object categories from different scenes, and is much larger than the preliminary version (\textbf{PAD}) \citep{Ours}. We provide rich affordance mask annotations, depth information annotations, bounding boxes annotations of humans/objects, and human pose annotations in support images. Besides, we train several representative models in related fields on PADv2 dataset and compare them with our OSAD-Net in terms of both objective evaluation metrics and visual quality for one-shot affordance detection. Our main contributions are summarised as follows:

\begin{itemize}
    \item [1)]
    We introduce a new challenging one-shot affordance detection problem along with a large-scale benchmark to facilitate the research for empowering robots with the ability to perceive unseen affordances in real-world scenarios.
    
    \item [2)]
    We propose a novel OSAD-Net that can efficiently learn the action purpose and use it to detect the common affordance of all objects in a scene via collaboration learning, resulting in a good adaptation capability that can deal with unseen affordances.
    
    \item [3)] 
    We establish a challenging PADv2 dataset containing $30$k images, covering $39$ affordance categories and $103$ object categories with more complex scenes. We provide rich annotations for the dataset, including pixel-level affordance mask labels, depth information annotations, bounding box annotations of human/objects and human pose annotations in support images, which could greatly benefit various visual affordance perception tasks.
    
    \item [4)]
    Experiments on the PADv2 and PAD datasets show that our OSAD-Net outperforms the state-of-the-art models and can serve as a strong baseline for future research.
\end{itemize}

\begin{table*}[t]
\caption{\textbf{Statistics of existing image-based affordance datasets and the proposed PADv2 dataset. PADv2 dataset provides higher-quality annotations and covers much richer affordance categories.} Pub.: Publication venue. Pixel: whether or not pixel-wise labels are provided. HQ: high-quality annotation. BG: the background is fixed or from general scenarios. $\sharp$Obj: number of object categories. $\sharp$Aff.: number of affordance categories. $\sharp$Img: number of images.}
\label{Table:x}
\scriptsize
\begin{center}
\renewcommand{\arraystretch}{1.}
\renewcommand{\tabcolsep}{7.5pt}
\begin{tabular}{c|c|cc|cccc|ccc}
\toprule[1.2pt] 
& Dataset     & Pub. & Year & Format & Pixel & HQ  & BG & $\sharp$Obj. & $\sharp$Aff. & $\sharp$Img.   \\
\midrule[1.2pt]
1 & 2011 ICRA \citep{hermans2011affordance}   & ICRA & 2011 & RGB-D  & -     & -   & Fixed      & -   & 7   & 375   \\
2 & UMD \citep{myers2015affordance}        & ICRA & 2015 & RGB-D  & \cmark   & \xmark  & Fixed      & -   & 17  & 30,000 \\
3 & 2016 TASE \citep{song2015learning}    & T-ASE & 2016 & RGB    & -     & -   & Fixed      & 8   & 1   & 10,360 \\
4 & IIT-AFF \citep{nguyen2017object}    & IROS & 2017 & RGB-D  & \cmark   & \xmark  & General    & 10  & 9   & 8,835  \\
5 & CERTH-SOR3D \citep{thermos2017deep} & CVPR & 2017 & RGB-D  & -     & -   & Fixed      & 14  & 13  & 20,800 \\
6 & ADE-Aff \citep{chuang2018learning}     & CVPR & 2018 & RGB    & \cmark   & \cmark & General    & 150 & 7   & 10,000 \\
\hline
7 & PAD \citep{Ours}    & IJCAI & 2021 & RGB    & \cmark   & \cmark & General    & 72  & 31  & 4,002  \\
\rowcolor{mygray}
8 & PADv2      &  &      & RGB-D  & \cmark   & \cmark & General    & 103  & 39  & 30,000 \\
\bottomrule[1.2pt]
\end{tabular}
\end{center}
\footnotesize{ 
$^2$\supp{\href{http://users.umiacs.umd.edu/~amyers/part-affordance-dataset/}{http://users.umiacs.umd.edu/~amyers/part-affordance-dataset/}}~~~$^4$\supp{\href{https://sites.google.com/site/ocnncrf/}{https://sites.google.com/site/ocnncrf/}} \\
	$^5$\supp{\href{http://sor3d.vcl.iti.gr/}{http://sor3d.vcl.iti.gr/}}~~$^6$\supp{\href{http://www.cs.utoronto.ca/~cychuang/learning2act/}{http://www.cs.utoronto.ca/~cychuang/learning2act/}}~~$^{7,8}$\supp{\href{https://github.com/lhc1224/OSAD_Net/}{https://github.com/lhc1224/OSAD\_Net/}}}
\end{table*}

A preliminary version of this work was presented in \cite{Ours}. In this paper, we extend the previous study by introducing three major improvements:

\begin{itemize}
    \item 
    We introduce a novel One-Shot Affordance Detection Network (OSAD-Net). Compared to OSAD-Net$^{\lozenge}$ $\footnote[2]{It indicates our conference version model in \cite{Ours}}$, we redesign all the three modules. Specifically, we consider the influence of human pose on learning action purpose, and introduce a probabilistic model for better purpose transfer and a dense comparison approach for feature enhancement of the same affordance object. In this way, we achieve better results than OSAD-Net$^{\lozenge}$ for one-shot affordance detection with fewer parameters and query images.
    
    \item 
    We extend the \textbf{P}urpose-driven \textbf{A}ffordance \textbf{D}ataset (PAD) further by collecting more images (up to $30$k), enlarging the diversity of the affordance category (up to $39$) and object category (up to $103$). We also provide posture annotations of support images and depth annotations of all images. Further, we evaluate more state-of-the-art methods from five related fields to comprehensively demonstrate the superiority of the proposed model.
    
    \item 
    We carefully re-organize the dataset and describe it in detail, including a complete statistical and attribute analysis, a clear definition of the problem, a comprehensive analysis of the experimental results from a variety of different perspectives, as well as complete ablation studies.
    
\end{itemize}

The remainder of the paper is organized as follows. Section \ref{related-work} describes existing works related to one-shot affordance detection. We introduce our OSAD-Net in Section \ref{Methodsection} and describe the benchmark dataset PADv2 in Section \ref{datasetsection}. Section \ref{experiments-section} presents the experimental results and analysis on both PADv2 and PAD datasets. We conclude the paper and discuss potential applications and future research directions in Section \ref{conclusionsection}.

\section{Related Work}
\label{related-work}
\subsection{Visual Affordance Learning}
Visual affordance is a branch of affordance research that deals with affordance as an image- or video-based computer vision problem and uses machine learning-related techniques to address the challenges \citep{hassanin2018visual}. In recent decades, a significant number of scholars try to explore object affordance from a vision perspective, which is divided into several main directions: affordance categorization, affordance detection, and affordance reasoning.
\par Affordance categorization is to predict the affordance category of an input image. \cite{stark2008functional} propose an algorithm to acquire, learn, and detect functional object categories based on affordance clues. They use human interaction videos to obtain a visual feature representation of affordance. \cite{ugur2014bootstrapping} use affordance cues of individual objects to guide the learning of complex affordance features of pairs of objects. Complex affordance learning is guided by using pre-learned basic visual features as additional input to complex affordance predictor variables or as cues to the next target object to be explored. The task of affordance detection is to divide the object into regions, and all pixels in each region are assigned with an affordance label. \cite{nguyen2017object} propose a deep learning-based object detector to improve affordance detection results, and subsequently, \cite{do2018affordancenet} improve this method and propose an end-to-end AffordanceNet. Unlike previous works that rely on separate and intermediate object detection steps, \cite{zhao2020object} propose a novel relationship-aware network to directly generate pixel-wise affordance maps from an input image in an end-to-end manner. In addition, to avoid extensive pixel-level annotation,  \cite{sawatzky2017weakly} and \cite{sawatzky2017adaptive} propose weakly supervised affordance detection methods, which can accomplish affordance detection using only a small number of keypoint annotations. Affordance reasoning refers to a more complex understanding of affordance, which requires higher-order contextual modeling, and the primary purpose of such reasoning is to infer hidden variables. \cite{zhu2014reasoning} build a knowledge base to represent target objects and their descriptive properties (visual, physical, and category properties) to infer affordance labels, human poses, or relative position, and learn the model using Markov logic networks \citep{richardson2006markov}. \cite{demo2vec2018cvpr} design the Demo2Vec model for extracting feature representations of demonstration videos and predicting the human interaction on the same target image regions and action labels on the same object image.

\par 
However, according to Gibson's definition of affordance \citep{gibson1977theory}, ``it implies the complementarity of the animal and the environment''. There exist multiple potential complementarities between animal and environment, which leads to multiple possibilities of particular affordance, \ie, an object may have multiple affordances, and the same affordance may cover multiple different object categories. This study attempts to establish a relationship between human action purpose and affordance and leverage a collaborative learning strategy to address the affordance ambiguity issue.

\subsection{Visual Affordance Dataset}

In the era of deep learning, the study of visual affordance usually follows a data-driven manner that requires annotated affordance datasets. \cite{hermans2011affordance} collect data from an autonomous mobile robot with a Pan-Tilt-Zoom (PTZ) camera, resulting in a total of $375$ images from $6$ object categories. \cite{myers2015affordance} introduce a large-scale RGB-D dataset containing pixel-level affordance labels and their ranks, which is the first pixel-wise labeled affordance dataset. \cite{song2015learning} propose a novel dataset for evaluating visual grasp affordance estimation. All images in the dataset contain grasp affordance annotations, including grasp region and scale attributes. Since most of the previous datasets have simple backgrounds and are difficult to be applied to real-world robot scenes due to their limited diversity, \cite{nguyen2017object} select a subset of object categories from ImageNet \citep{russakovsky2015imagenet} and collect RGB-D images from clutter scenes for constructing the IIT-AFF dataset. \cite{thermos2017deep} propose an RGB-D sensorimotor dataset for the sensorimotor object recognition task. However, the affordances of objects do not simply correspond to appearance, which are shifted in response to the state of interactions between objects and humans. Therefore, \cite{chuang2018learning} consider the problem of affordance reasoning in the real-world by taking into account both the physical world and the social norms imposed by the society and constructed the ADE-Affordance dataset based on ADE20k \citep{zhou2017scene}. Different from these works, \cite{Ours} construct a Purpose-driven Affordance Dataset (PAD) considering the relationship between human purpose and affordance, which involves more complex scenarios and thus potentially benefits practical robot applications.

\par In addition to the image-based affordance datasets presented above, some datasets consider other aspects of affordance. \cite{wang2017binge} extract a diverse set of scenes and how actors interact with different objects in the scenes from seven sitcoms. Subsequently, \cite{li2019putting} extend this work to 3D indoor scenes and construct a 3D pose synthesizer that fuses semantic knowledge from 2D poses extracted from TV shows as well as 3D geometric knowledge from voxel representations of indoor scenes. Recently, \cite{deng20213d} propose a 3D AffordanceNet dataset containing $23$k shapes from $23$ semantic object categories annotated with $18$ visual affordance categories.

\par In this paper, we focus on the visual affordance detection task and try to construct a benchmark to facilitate the research in this area. Specifically, we expand the size and diversity of the Purpose-driven Affordance Dataset in our preliminary work \citep{Ours} and establish a large-scale PADv2 dataset with more complex scenes and richer affordance and object categories. Our dataset contains pixel-level and image-level labels and the depth information of the image to provide more comprehensive information for future study of affordance detection. Statistics about the existing image-based affordance datasets are summarized in Table \ref{Table:x}.

\subsection{One-Shot Learning}
Few-shot learning refers to learning a model that can recognize new sample classes given a few of reference images, which primarily concerns about the model's generalization ability. Existing works focus on metric-based, meta-based, and augmentation-based methods. The core idea of metric-based methods \citep{cai2014attribute,snell2017prototypical,sung2018learning,vinyals2016matching} is to optimize the distance/similarity between images or regions. While the meta-based approaches \citep{finn2017model,he2020progressive,rusu2018meta,zhu2020self,ravi2016optimization,wang2016learning} mainly define a specific objective or loss function to guide the model to get a fast learning capability. Augmentation-based approaches \citep{li2020adversarial} mainly consider synthesizing more data in different ways from new categories to facilitate the learning stage.

\par Few-shot segmentation is a more challenging task to predict a label for each pixel instead of one for the whole image. \cite{shaban2017one} propose a typical two-branch network. Later, \cite{dong2018few} introduce the idea of prototype. \cite{zhu2019one} propose an one-shot texture retrieval (OS-TR) network. Given an example of a new reference texture, the network detects and segments all the pixels of the same texture category within an arbitrary image. CANet \citep{zhang2019canet} averages the object features in the support image, extends them to the size of the query feature and then concatenates them together, and leverages an iterative optimization module to refine the segmentation results. To cope with the problem of representing prototypes with ambiguity and intra-class variations due to the lack of training samples, \cite{wang2021variational} leverage probabilistic hidden variables to represent prototype distributions, converting the discriminative model into a probabilistic model that allows for a more expressive representation of object categorical concepts. In addition, they represent optimization as a variational inference problem. \cite{johnander2021deep} propose a few-shot learner formulation based on gaussian process regression, enabling the network to model complex object apparent distributions in deep feature space. \cite{li2021adaptive} utilize superpixel-guided clustering to generate multiple prototypes and allocate them to query features.

\par Our one-shot affordance detection task is quite different from one-shot segmentation. (1) One-shot affordance detection focuses on the affordance property of the object, which is not completely equivalent to its semantics, since the same object may have multiple affordances and different objects may also share the same affordance. (2) The inputs required for the two tasks are different. One-shot segmentation requires a support image and a pixel-wise object mask, while we only need the bounding boxes of the human and object and the human pose in the one-shot affordance detection task, which are easier to obtain by off-the-shelf object detectors and human pose detection networks.

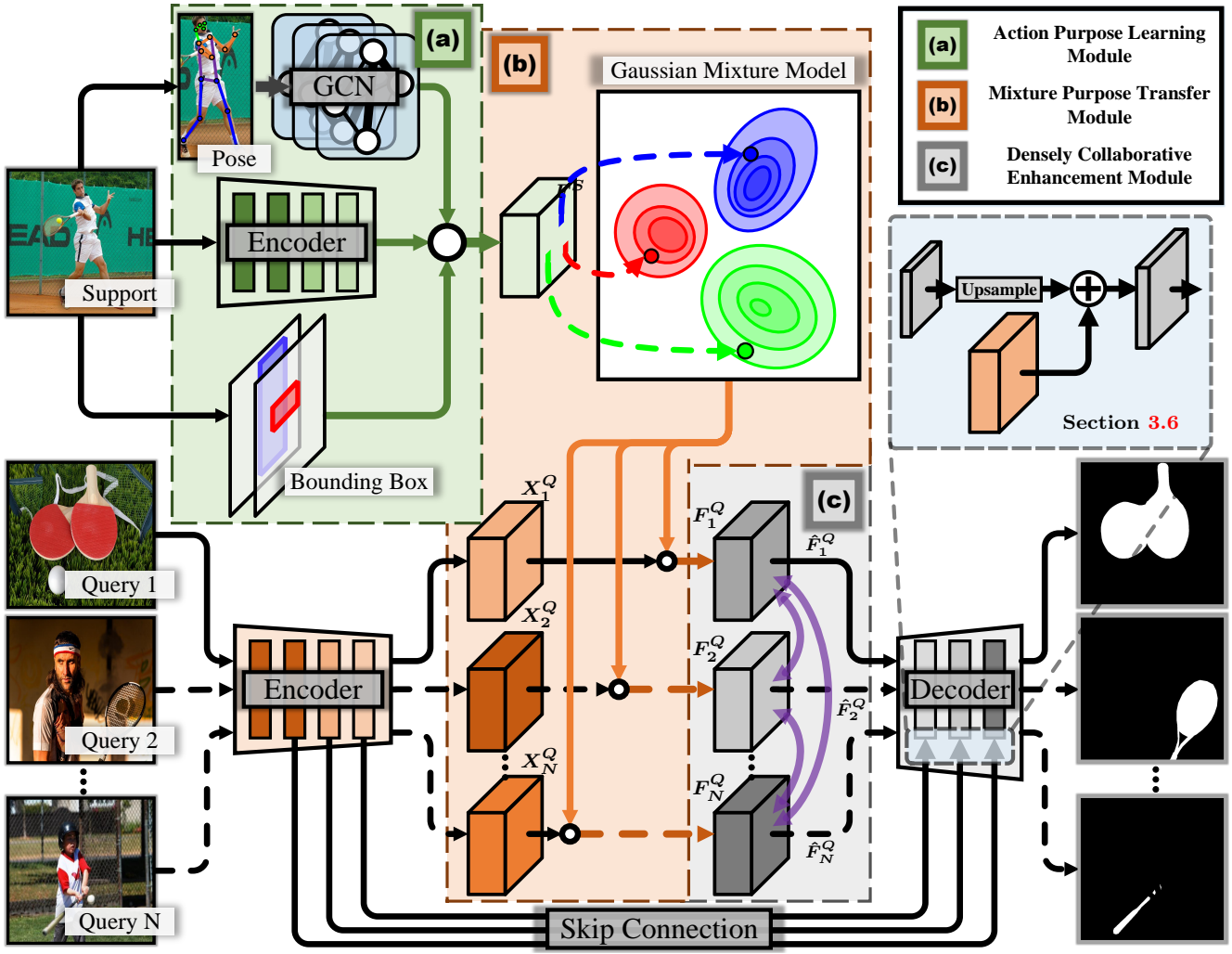
\begin{figure*}[t]
	\centering
	\begin{overpic}[width=1.\linewidth]{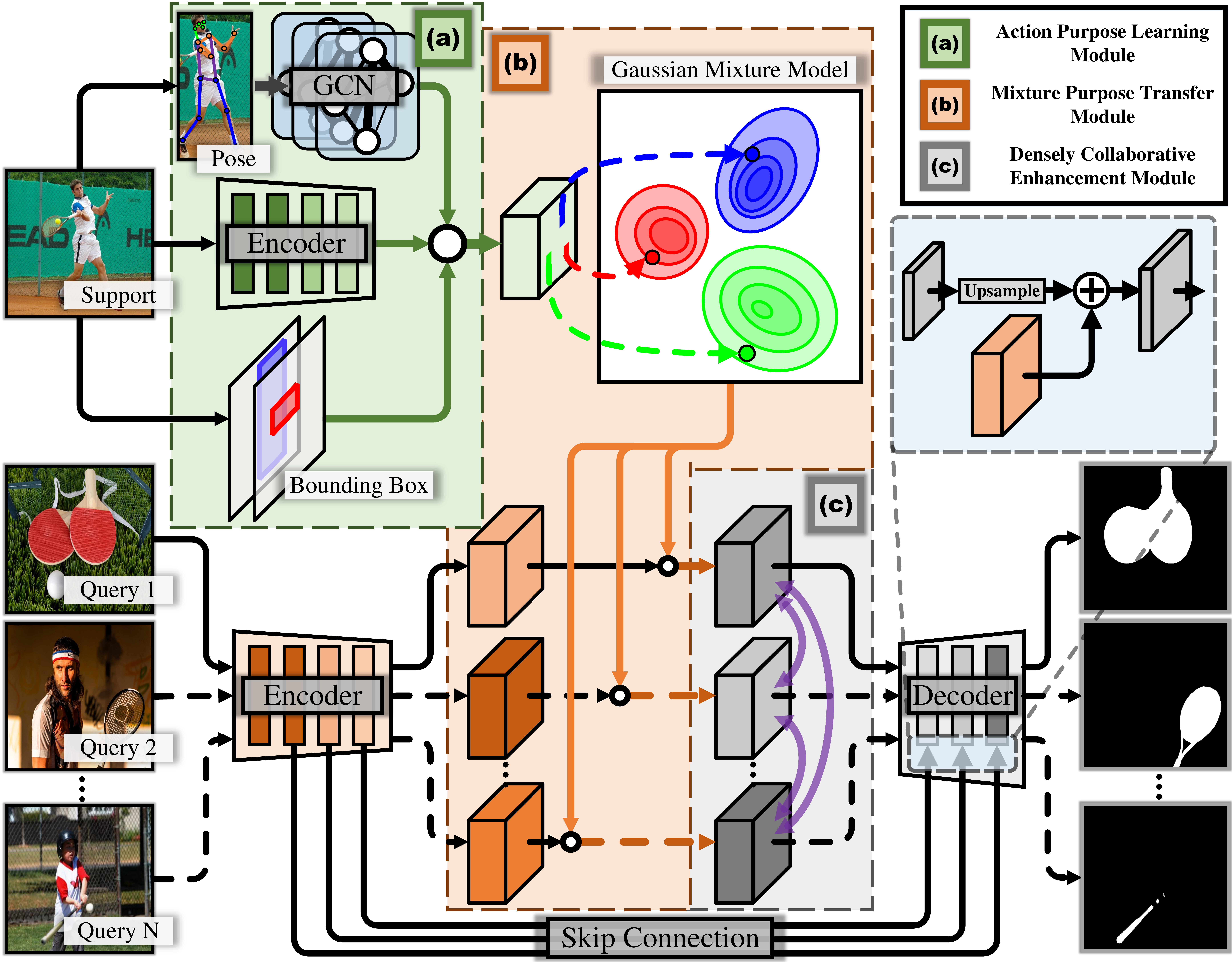}
        \put(45,62){\small{$\bm{F^S}$}}
        \put(42.3,37.6){\small{$\bm{X^Q_1}$}}
        \put(42.3,27.2){\small{$\bm{X^Q_2}$}}
        \put(42.3,15.2){\small{$\bm{X^Q_N}$}}
        \put(56.3,35.4){\small{$\bm{F^Q_1}$}}
        \put(56.3,24.4){\small{$\bm{F^Q_2}$}}
        \put(56.3,13.0){\small{$\bm{F^Q_N}$}}
        \put(65.5,33.0){\scriptsize{$\bm{\hat{F}^Q_1}$}}
        \put(68.0,19.5){\scriptsize{$\bm{\hat{F}^Q_2}$}}
        \put(65.5,7.8){\scriptsize{$\bm{\hat{F}^Q_N}$}}
        \put(86.5,43){\small{\textbf{Section \ref{decoder section}}}}
    \end{overpic}
	\caption{\textbf{The framework of the proposed OSAD-Net.} OSAD-Net first uses a Resnet50 \citep{he2016deep} to extract the features of support image and query images. Subsequently, the support feature, the bounding boxes of the person and object, and the pose of the person are fed into the APL module (see Section \ref{plm} for details) to obtain the human action purpose feature. Next, the human action purpose feature and query image feature are sent to the MPT module (see Section \ref{ptm} for details) to transfer the human action purpose to query images and activate the object regions having the target affordance in the query images. Then, the output of the MPT is fed into a DCE module (see Section \ref{cem}) to learn the commonality among objects of the same affordance and suppress the irrelevant background regions using a collaborative learning strategy. Finally, the enhanced features are fed into the decoder (see Section \ref{decoder section}) to obtain the detection results.}
	\label{pipeline}
\end{figure*}

\section{Method}
\label{Methodsection}
\subsection{Problem Description}
The one-shot affordance detection task consists of two main sets, \ie, the query set $Q$ and support set $S$. Given a support image from $S$ and a set of query images from $Q$, the goal of the task is to segment all objects in the query images with the same affordance based on the information provided by the support image. Models are trained on classes $C_{train}$ (base) and tested on previously unseen class $C_{test}$ (novel) in episodes ($C_{train} \cap C_{test}= \emptyset$). Each episode is formed by a support set $S$ and a query set $Q$ of the same class $c$. The support set $S$ only contains one sample $S=\{I^S, B^h, B^o, P\}$ of class $c$, where $I^S$ is the support image, $B^h$ is the human bounding box, $B^o$ is the object bounding box, and $P$ is the human pose. For the query set, we define it as $Q=\{(I^{Q}_1, Y_1), (I^{Q}_2, Y_2), ..., (I^{Q}_N, Y_N)\}$, where $I^Q_i$ is the $i$-th query image and $Y_i$ is its affordance mask label, $N$ is the number of query samples. In each batch, we define the input as a support sample and $N$ query images.

\subsection{Pipeline}
\begin{figure}[t]
	\centering
	\begin{overpic}[width=0.99\linewidth]{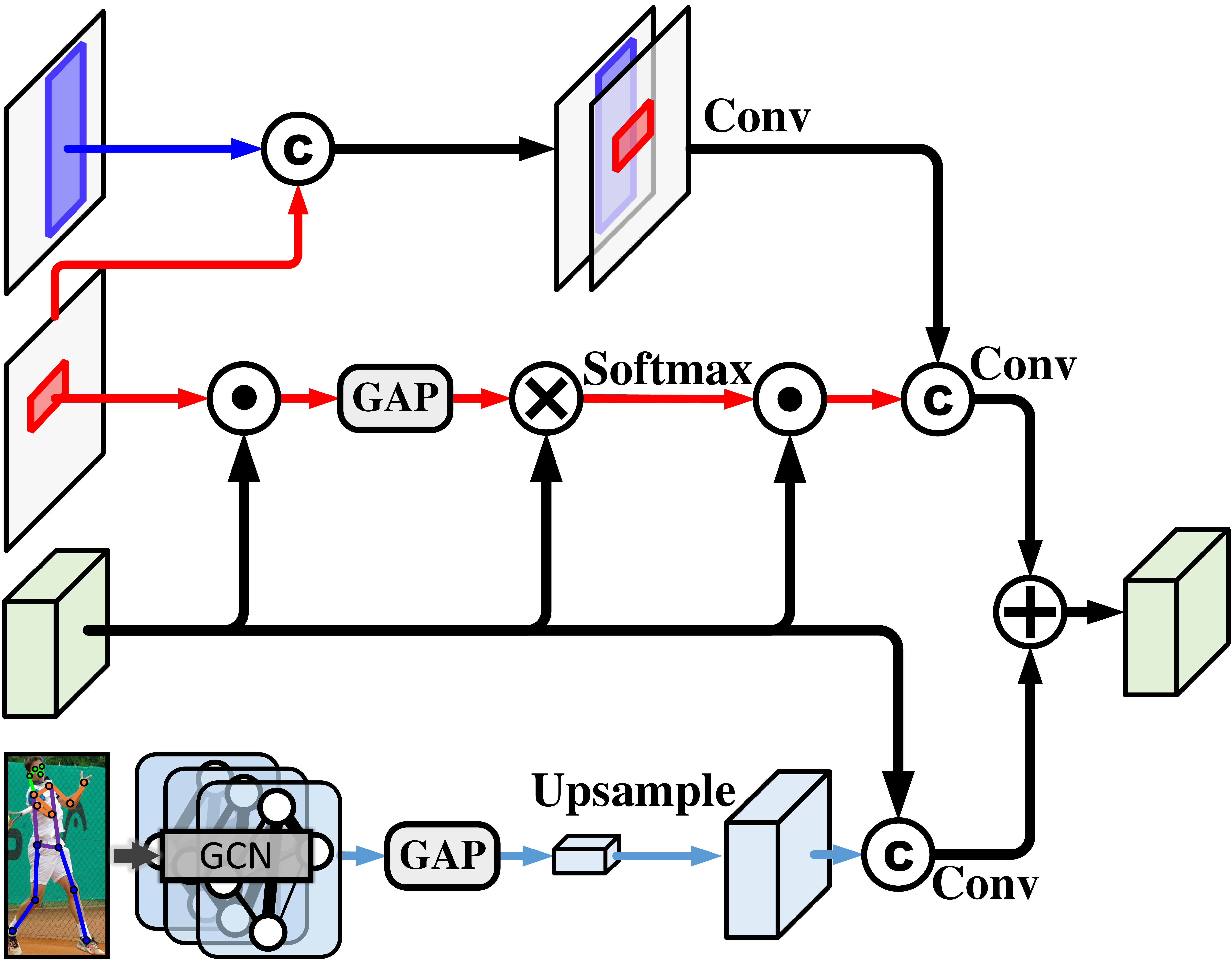}
        \put(10,72.3){\footnotesize{$\bm{B^h}$}}
        \put(10,50){\footnotesize{$\bm{B^o}$}}
        \put(10,30){\footnotesize{$\bm{X^S}$}}
        \put(9,18){\footnotesize{$\bm{P}$}}
        \put(45,3){\footnotesize{$\bm{\tilde{P}}$}}
        \put(84.5,16){\footnotesize{$\bm{F^P}$}}
        \put(68,47){\footnotesize{$\bm{F^o}$}}
        \put(77,55.5){\footnotesize{$\bm{S^{ho}}$}}
        \put(84.5,38.5){\footnotesize{$\bm{F^{ho}}$}}
        \put(95.3,37){\footnotesize{$\bm{F^S}$}}
        \put(38,48){\footnotesize{$\bm{f^o}$}}
        \put(21,49){\footnotesize{\textbf{Conv}}}
    \end{overpic}
	\caption{\textbf{The action purpose learning module}. It mainly considers the object appearance, the relative position relationship between the human and object, and the human's pose to reason about the action purpose jointly.}
	\label{Module-1}
\end{figure}
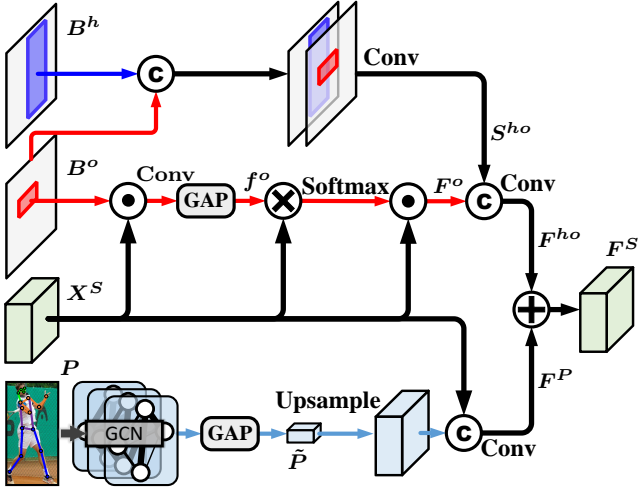

\begin{figure*}[t]
	\centering
	\begin{overpic}[width=0.94\linewidth]{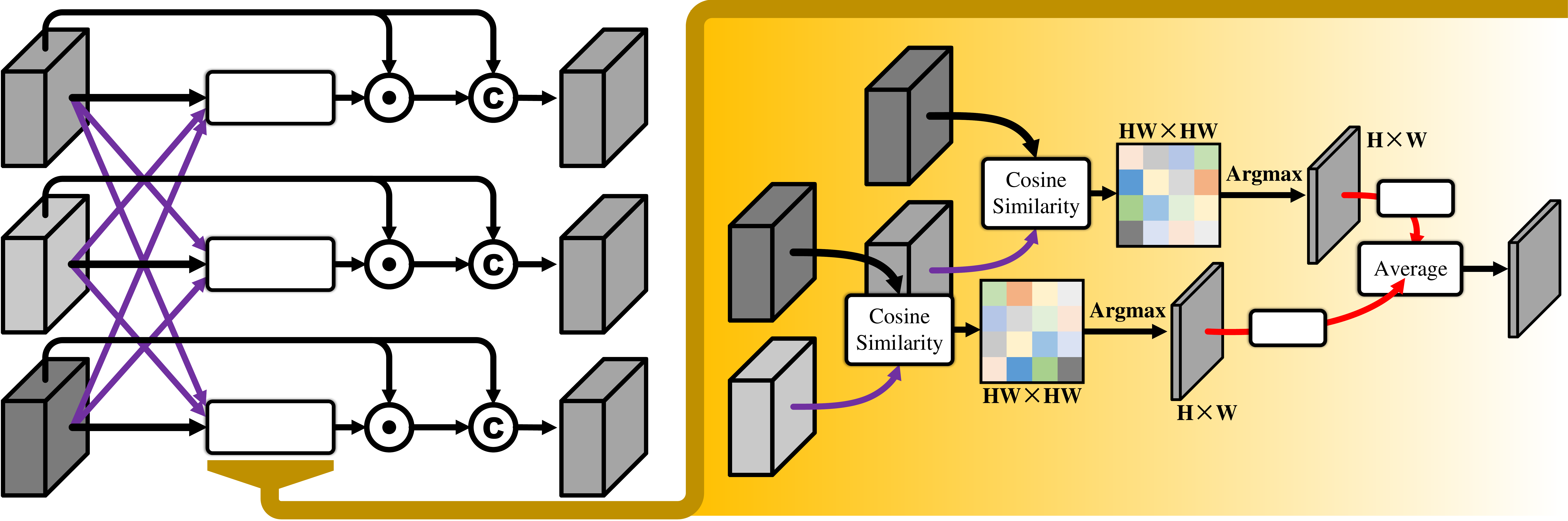}
        \put(13.4,16){\footnotesize{\textbf{Eq.\ref{eq:9}$\sim$\ref{eq:13}}}}
        \put(13.4,26.5){\footnotesize{\textbf{Eq.\ref{eq:9}$\sim$\ref{eq:13}}}}
        \put(13.4,5.5){\footnotesize{\textbf{Eq.\ref{eq:9}$\sim$\ref{eq:13}}}}
        \put(-3,25.5){\footnotesize{$\bm{F^Q_1}$}}
        \put(-3,15){\footnotesize{$\bm{F^Q_2}$}}
        \put(-3,4.5){\footnotesize{$\bm{F^Q_N}$}}
        \put(41.5,27){\footnotesize{$\bm{\hat{F}^Q_1}$}}
        \put(41.5,16.5){\footnotesize{$\bm{\hat{F}^Q_2}$}}
        \put(41.5,6.5){\footnotesize{$\bm{\hat{F}^Q_N}$}}
        \put(52,5){\footnotesize{$\bm{F^Q_2}$}}
        \put(51.6,14.5){\footnotesize{$\bm{F^Q_N}$}}
        \put(61,17){\footnotesize{$\bm{F^Q_1}$}}
        \put(61,27){\footnotesize{$\bm{F^Q_N}$}}
        \put(79.6,11.9){\scriptsize{\textbf{Eq.\ref{eq:12}}}}
        \put(87.8,20.1){\scriptsize{\textbf{Eq.\ref{eq:12}}}}
        \put(85,11.5){\footnotesize{$\bm{M^Q_{N2}}$}}
        \put(91.5,18.2){\scriptsize{$\bm{M^Q_{N1}}$}}
        \put(96,10){\footnotesize{$\bm{\bar{M}_{N}^Q}$}}
        \put(21,13.2){\scriptsize{$\bm{\bar{M}_{2}^Q}$}}
        \put(21,2.8){\scriptsize{$\bm{\bar{M}_{N}^Q}$}}
        \put(21,24){\scriptsize{$\bm{\bar{M}_{1}^Q}$}}
    \end{overpic}
	\caption{\textbf{The densely collaborative enhancement module.} We perform a dense comparison between two query images and calculate their correlation, so as to suppress the background unrelated to affordance and obtain better segmentation results.}
	\label{Module-3}
\end{figure*}
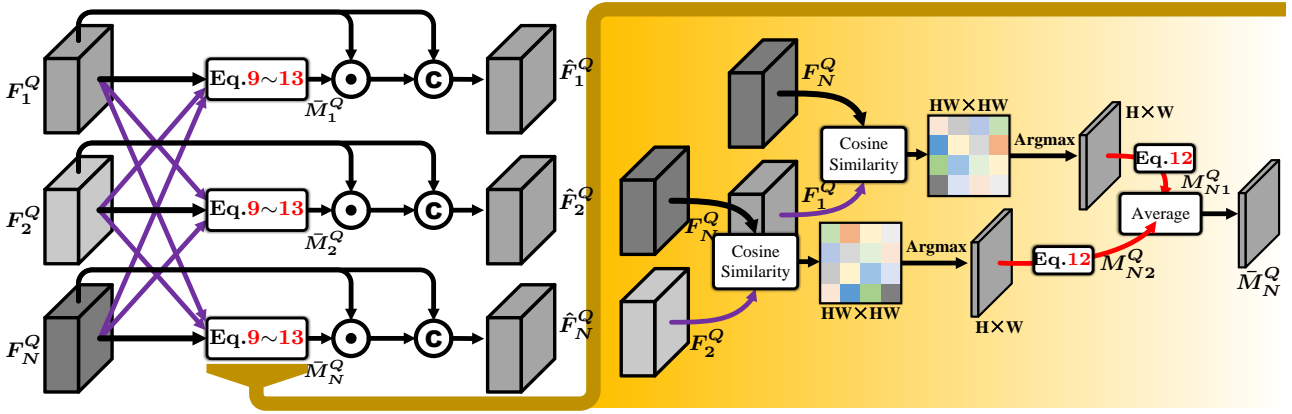

In this section, we briefly introduce the \textbf{O}ne-\textbf{S}hot \textbf{A}ffordance \textbf{D}etection \textbf{Net}work (OSAD-Net), as shown in Fig. \ref{pipeline}. We first feed both the image $I^S$ in the support set $S$ and the images $I^{Q}_i  (i=1,2,...,N)$ in the query set $Q$ into a backbone network to extract features. In this paper, we use Resnet50 \citep{he2016deep} as the backbone. Then, the feature maps of support image, the bounding boxes of human and object, and the pose of human are fed into the APL module (as shown in Fig. \ref{Module-1}) to estimate the human action purpose feature, which implicitly defines the affordance in the current state. By leveraging the human-object interaction and the human pose information, APL can mitigate the affordance ambiguity issue brought by the fact that an object may have multiple affordances, \ie, multiple affordance possibilities collapse to an explicit affordance given the action purpose. Subsequently, we input the action purpose feature and the query image features into the MPT module to transfer the human action purpose into query images and activate the object regions belonging to the same affordance. After that, we feed the features of the activated query images into a DCE module (as shown in Fig. \ref{Module-3}) to obtain enhanced features by mining the commonality among objects of the same affordance and effectively eliminate the influence of the appearance differences between object classes. Finally, the enhanced features are fed into the decoder to obtain the final detection results. 

\subsection{Action Purpose Learning}
\label{plm}
The goal of the APL module is to infer the purpose of human action from the support image. To this end, we try to discover the human action purpose from three clues: human pose, the relative position of human and object, and the properties of object appearance  \citep{xu2019interact,zhong2021polysemy,wei2017inferring}. For example, the bicycle wheel allows us to associate it with ``Push'', while the structure of the bicycle provides the affordance of ``Ride'', and the relative position of the person and the object, \eg, up and down or left and right, can be used to determine whether it is a bicycle or a cart. Besides, the human pose also provides essential clues for reasoning about human action purpose, \eg, the fact that a person's leg is bent or straight can determine whether the action is cycling or pushing. 

\par As shown in Fig. \ref{Module-1}, the APL module receives four inputs including the support feature $X^S$, the bounding box $B^h$ of the human, the bounding box $B^o$ of the object, and the pose $P$ of the human. Inspired by \cite{kipf2016semi,yan2018spatial}, we use the Graph Convolutional Network (GCN) to process the human pose $P$, which is defined as follows:
\begin{equation}
    GCN(P)=\Lambda^{-\frac{1}{2}}(A+I)\Lambda^{-\frac{1}{2}}PW, \label{eq:1}
\end{equation}
where $P$ is the coordinates of the keypoints' positions, $\Lambda_{ii}=\sum_j(A_{ij}+I_{ij})$, and $W$ is the learnable weight matrix. Meanwhile, we introduce a learnable weight matrix $M$, which is multiplied with $A$ to measure the importance of edges. $M$ is initialized as an all-ones matrix. Therefore, the GCN operation defined in this paper is as follows:
\begin{equation}
    GCN(P)=\Lambda^{-\frac{1}{2}}(A \otimes M+I)\Lambda^{-\frac{1}{2}}PW. \label{eq:2}
\end{equation}
We feed the skeletal data through four layers of GCN and add a residual connection to the output of each layer. Afterward, we leverage a global average pooling layer after the final layer of GCN to obtain the output of the skeleton branch $\tilde{P}$. Then, we expand $\tilde{P}$ to the same size as $X^S$, concatenate it with $X^S$, and pass them through a convolution layer to obtain the relevant region feature $F^P$ activated by the intrinsic linkage provided by pose, as illustrated in Fig. \ref{Module-1}.

\begin{figure}[t]
	\centering
	\begin{overpic}[width=0.95\linewidth]{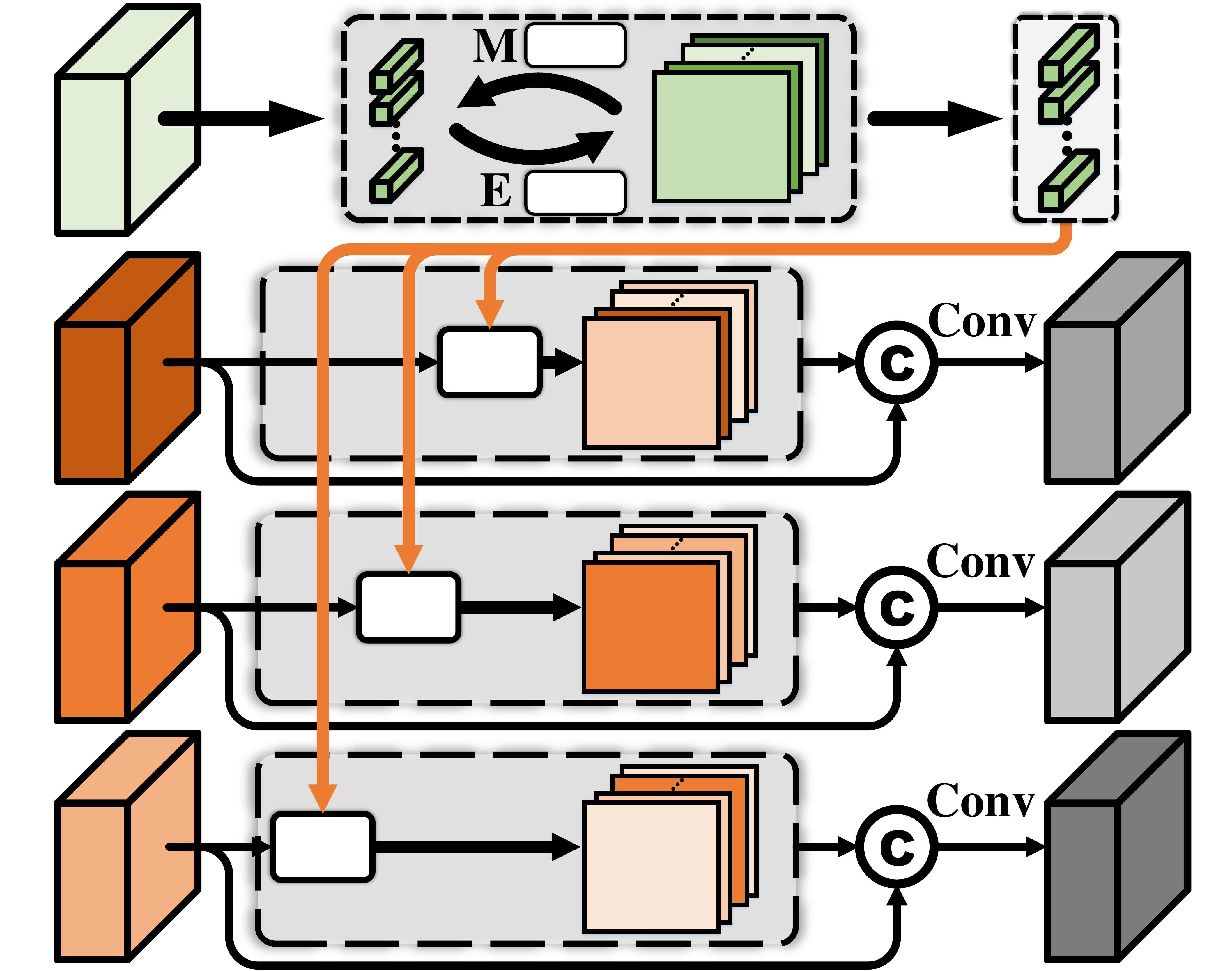}
        \put(42.65,62.3){\scriptsize{\textbf{Eq.\ref{eq:6}}}}
        \put(42.65,74.2){\scriptsize{\textbf{Eq.\ref{eq:7}}}}
        \put(22,9.3){\scriptsize{\textbf{Eq.\ref{eq:8}}}}
        \put(29,28.7){\scriptsize{\textbf{Eq.\ref{eq:8}}}}
        \put(35.7,48.5){\scriptsize{\textbf{Eq.\ref{eq:8}}}}
        \put(-1,66){\scriptsize{$\bm{F^S}$}}
        \put(-2,47){\scriptsize{$\bm{X^Q_1}$}}
        \put(-2,27){\scriptsize{$\bm{X^Q_2}$}}
        \put(-2,7){\scriptsize{$\bm{X^Q_N}$}}
        \put(32,79){\scriptsize{$\bm{\mu}$}}
        \put(60,78.5){\scriptsize{$\bm{Z^S}$}}
        \put(50,47){\scriptsize{$\bm{Z^Q_1}$}}
        \put(50,27){\scriptsize{$\bm{Z^Q_2}$}}
        \put(50,7){\scriptsize{$\bm{Z^Q_N}$}}
        \put(65,53){\scriptsize{$\bm{\tilde{X}^Q_1}$}}
        \put(65,33){\scriptsize{$\bm{\tilde{X}^Q_2}$}}
        \put(65,13.5){\scriptsize{$\bm{\tilde{X}^Q_N}$}}
        \put(97,49.5){\scriptsize{$\bm{F^Q_1}$}}
        \put(97,30){\scriptsize{$\bm{F^Q_2}$}}
        \put(97,10){\scriptsize{$\bm{F^Q_N}$}}
    \end{overpic}
	\caption{\textbf{The mixture purpose transfer module.} Transferring the human action purpose to query images implies to activate object regions with the same affordance, which can be modeled as a gaussian mixture model and the action purpose can be represented in a form of a compact set of bases.}
	\label{Module-2}
\end{figure}

\par At the same time, we leverage object features for action purpose inference, \ie, using the object features to activate the relevant regions of human-object interaction. As shown in Fig. \ref{Module-1}, we first perform a dot product operation between the object bounding box and support feature to mask the object, which is then fed into a convolution layer and a global average pooling layer to obtain the feature representation of the object $f^o$. Next, we calculate the correlation coefficient between $f^o$ and each location in $X^S$. After normalizing the correlation coefficients using a Softmax function, we obtain the attention weights, which are then multiplied with $X^S$ using dot product to obtain the output $F^o$. The calculation process can be formulated as follows:
\begin{equation}
   F^o=Softmax(X^S \otimes f^o) \odot X^S. \label{eq:3}
\end{equation}

\par The relative position of the person and the object also provide critical information for reasoning about the human action purpose. Therefore, we concatenate the bounding boxes of the person and object, feed them into a convolution layer to obtain the relative position feature $S^{ho}$, and then concatenate it with $F^o$. The final human-object interaction representation $F^o$ is obtained after a convolution layer:
\begin{align}
    S^{ho}=Conv(Concat(B^o,B^h)), \label{eq:4} \\
    F^{ho}=Conv(concat(F^o,S^{ho})). \label{eq:5}
\end{align}
\par Finally, we sum $F^p$ and $F^{ho}$ to get $F^{S}$, \ie, the feature representation of human action purpose.

\subsection{Mixture Purpose Transfer}
\label{ptm}
After inferring the action purpose from the support image by the APL module, we introduce the MPT module to transfer the action purpose to the query images, activating all object regions that can accomplish that purpose. Since different categories of objects may have the same affordance, there are significant differences in appearance between the objects. In addition, there are great variations in poses for the same action purpose, therefore bringing a challenge to the transfer process. Inspired by the probabilistic model \citep{johnander2021deep,wang2021variational,dempster1977maximum,sun2021attentional}, we use a Gaussian mixture model \citep{richardson1997bayesian} to account for the differences within the same affordance category and use multiple affordance-related purpose prototypes to jointly represent an action purpose. As shown in Fig. \ref{Module-2}, we use a compact set of bases $\mu$ to encode the action purpose, which can be obtained by Expectation-Maximization (EM) iteration \citep{li2019expectation}.

\par In the MPT module, we run ``E-step'' and ``M-step'' alternately on $F_S$ to obtain a compact set of bases and then use them to reconstruct the query features. For the input $F^S$ ($F^S \in R^{L \times C}, L=H \times W $), a set of bases $\mu \in R^{K \times C}$ is firstly randomly initialized. Then, the E-step process estimates the latent variable $Z^S$ ($Z^S \in R^{L \times K}$) from $F^S$. The weight of the $k$-th base at the $j$-th pixel on $F^S$ is calculated as follows:
\begin{equation}
    Z^S_{jk}=\frac{\kappa(f^S_{j},\mu_k)}{\sum_{l=1}^K\kappa(f^S_j,\mu_l)}, \label{eq:6}
\end{equation}
where $f^S_j$ is the feature at the $j$-th position of $F^S$ and $\kappa$ is defined as the exponential kernel function, \ie, $\exp(\cdot)$. Thus, we can compute $Z^S = Softmax(F^S(\mu)^T)$. In the M-step, $\mu$ is computed as a weighted average of $F^S$. Concretely, the $k$-th basis is updated as:
\begin{equation}
    \mu_k=\frac{\sum_{j=1}^L Z^S_{jk}f^S_j}{\sum_{j=1}^L Z^S_{jk}}. \label{eq:7}
\end{equation}
After several EM iterations, we obtained a set of bases, which is then used to reconstruct the query image features. For the input set of query image features $\mathcal{X^Q}=\{X^{Q}_1, ..., X^{Q}_N\}$, we calculate the attention weights $\mathcal{Z^Q}=\{Z^{Q}_1, ..., Z^{Q}_N\}$ as:
\begin{equation}
    Z^{Q}_i=Softmax(X^{Q}_i(\mu)^T). \label{eq:8}
\end{equation}
Then, we use $\mu$ and $\mathcal{Z^Q}$ to reconstruct the query image feature as $\mathcal{\tilde{X}^Q}$, \ie, $\tilde{X}^{Q}_i=Z^{Q}_i\mu$. Finally, we concatenate $\mathcal{X^Q}$ and $\mathcal{\tilde{X}^Q}$ to obtain the output $\mathcal{F^Q}=\{F^Q_1,...,F^Q_N\}$ of the MPT module after a convolution layer.

\subsection{Densely Collaborative Enhancement Module}
\label{cem}
After transferring the human action purpose from the support image to the query images, the object regions activated by the MPT module are coarse due to the limited representation ability of purpose feature given the fact that different categories of objects may have the same affordance and they have different appearances. In order to obtain more accurate affordance object regions, we introduce a DCE module to enhance the feature representation by capturing the common relationship between the objects having the same affordance.
\par As shown in Fig. \ref{Module-3}, for the query feature $F_Q^i$, we compute the correlation mask $M^Q_{ij}$ between $F^Q_i$ and $F^Q_j$ ($j \in [1,N], j\ne i$) by dense comparison. We first compute the pixel-level cosine similarity $cos(f_k,f_l)$ between $f_k \in F^Q_i$ and $f_l \in F^Q_j$ as follows:
\begin{align}
  cos(f_k,f_l)= \frac{(f_k)^T f_l}{\lVert f_k\lVert \lVert f_l \lVert}, \label{eq:9}
\end{align}
For each $f_k \in F^Q_i$, we take the maximum similarity among all $F^Q_j$ pixels as the correspondence value:
\begin{align}
	&c_{ijk}=\mathop{\max}\limits_{l\in \{1,2,...,HW\}}(cos(f_k,f_l)), \label{eq:10} \\
	&C^Q_{ij}=[c_1,c_2,...,c_{HW}] \in R^{HW \times 1}.\label{eq:11}
\end{align}
Then, we reshape $C^Q_{ij}$ from $R^{HW \times 1}$ to $R^{H\times W \times 1}$ and perform min-max normalization to obtain the correlation mask $M^Q_{ij}$ of $F^Q_i$ for query $F^Q_j$. For query $F^Q_i$, $N-1$ correlation masks can be obtained in this way. Then we calculate the average of $N-1$ correlation masks to obtain the output $\bar{M}^Q_i$:
\begin{align}
    &M^Q_{ij}=\frac{C^Q-min(C^Q)}{max(C^Q)-min(C^Q)}, \label{eq:12} \\
    &\bar{M}^Q_i=\frac{1}{N-1}\sum_{j}^{N-1} M^Q_{ij}. \label{eq:13}
\end{align}
Finally, $\bar{M}^Q_i$ is multiplied with $F^Q_i$ to obtain the enhanced feature, which is then concatenated with $F^Q_i$ and fed into a convolution layer to get the $i$-th query feature $\hat{F}^Q_i$.

\begin{figure*}[t]
	\centering
	\begin{minipage}[c]{1.\linewidth}
    \scriptsize
    \setlength{\fboxrule}{1.5pt}
	\begin{overpic}[width=1.\linewidth]{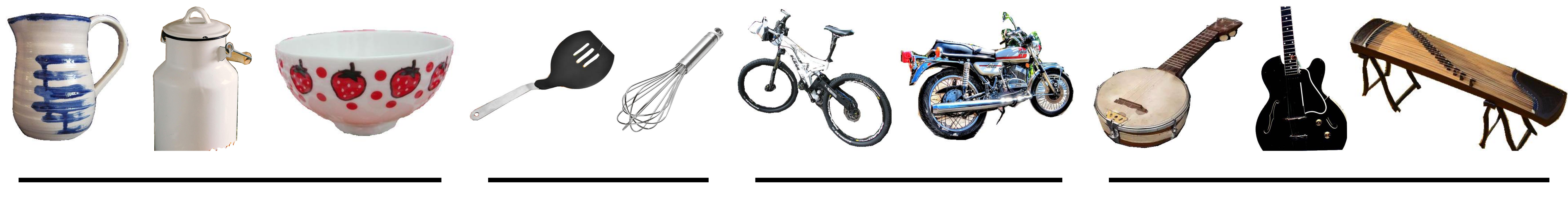}
           \put(1.5,2.3){pitcher}
           \put(9.5,2.3){milk can}
           \put(21.5,2.3){bowl}
           \put(11,-0.5){\small\textbf{Contain-2}}
           
           \put(32,2.3){spatula}
           \put(40.5,2.3){whisk}
           \put(36.5,-0.5){\small\textbf{Mix}}
           
           \put(49.5,2.3){bicycle}
           \put(60,2.3){motorbike}
           \put(56.5,-0.5){\small\textbf{Ride}}
            
           \put(71.5,2.3){banjo}
           \put(80.5,2.3){guitar}
           \put(91,2.3){zither}
           \put(82,-0.5){\small\textbf{Play-2}}
    \end{overpic}
    \end{minipage}
	\caption{\textbf{Some examples from our PADv2 dataset.} We visualize the foreground objects from some representative affordance categories.}
	\label{dataset_ins}
\end{figure*}

\begin{figure*}[t]
	\centering
	\begin{minipage}[c]{1.\linewidth}
    \centering
    \setlength{\fboxrule}{1.5pt}
	\begin{overpic}[width=1.\linewidth]{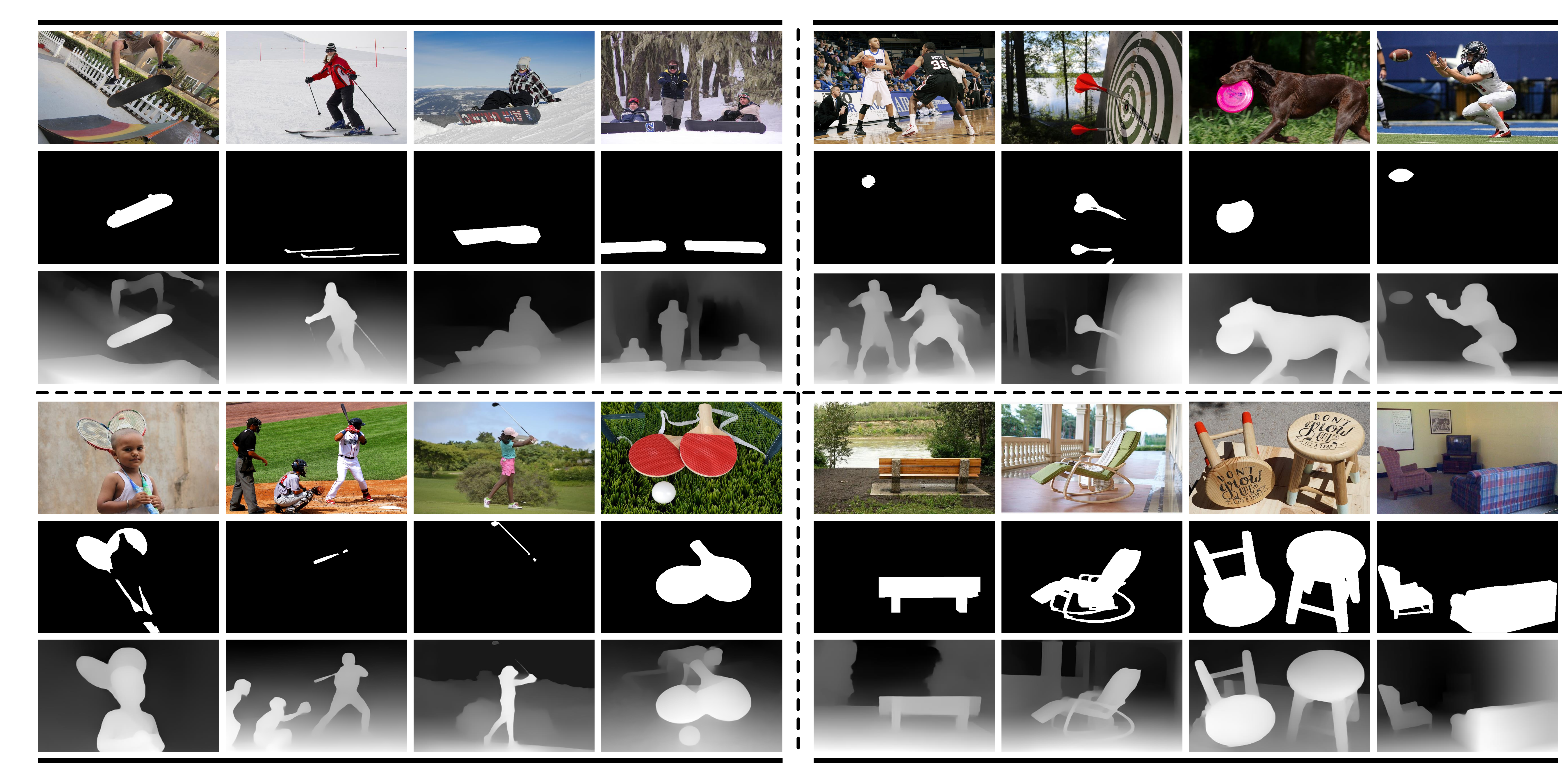}
           \put(23,48){\fcolorbox{black}{white}{Jump}}
           \put(72,48){\fcolorbox{black}{white}{Throw}}
           \put(22.5,1.5){\fcolorbox{black}{white}{Swing}}
           \put(74,1){\fcolorbox{black}{white}{Sit}}
    \end{overpic}
    \end{minipage}
	\caption{\textbf{Images and their annotations from our PADv2 dataset.} PADv2 dataset has rich annotations such as affordance masks (the second row in each group) as well as depth information (the third row in each group).}
	\label{dataset example}
\end{figure*}
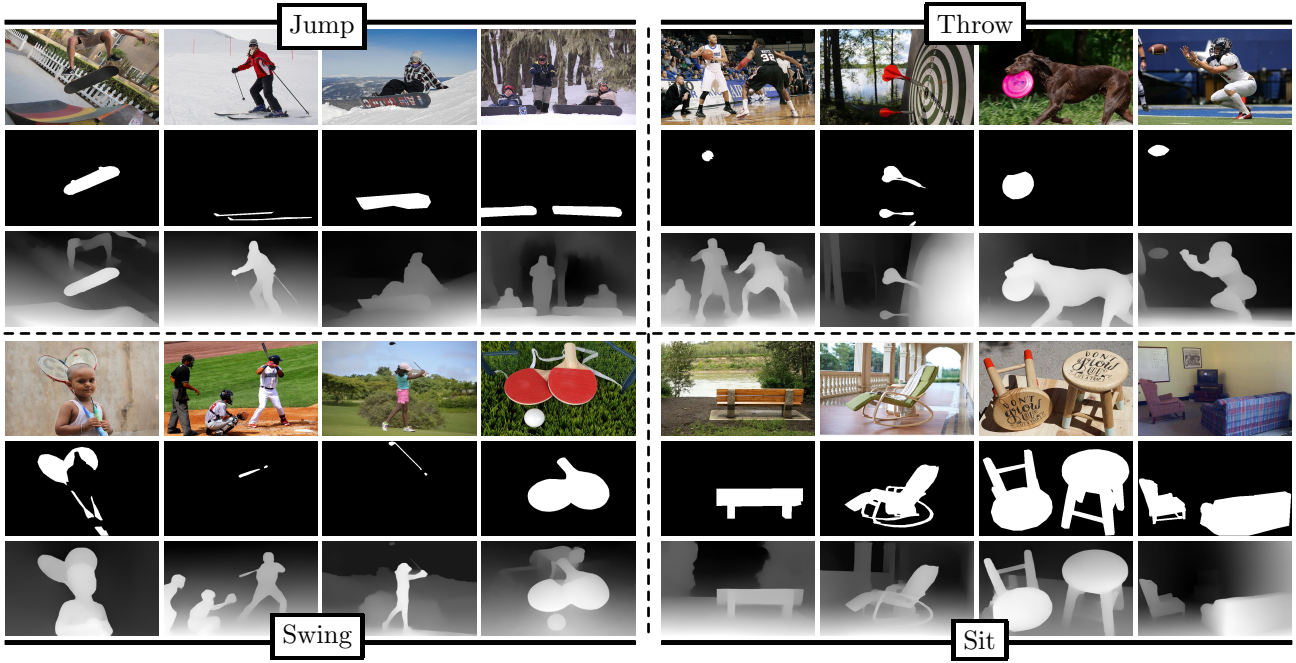

\subsection{Decoder}
\label{decoder section}
For the $i$-$th$ query image, the output of the $m$-$th$ decoder layer is denoted as $P^Q_{i, m}=\text{Conv}(\text{Conv}(X^{Q}_{i, m})+\text{Upsample}(P^{Q}_{i, m+1}))$, $m\in [1, 4]$, where $P^{Q}_{i, 5}=\text{Conv}(\hat{F}^{Q}_{i})$. Subsequently, a convolutional prediction layer is appended to each $P^Q_{i, m}$ to get the side output, \ie, $D^{Q}_{i, m}=\text{Conv}(P^{Q}_{i, m})$, $m\in [1, 5]$. The cross-entropy loss is used as the training objective. For $D^Q_{i,m}$, we calculate the loss $\mathcal{L}_{i,m}$ as: 
\begin{align}
     \mathcal{L}_{i,m}=-\frac{1}{H\times W}&(\sum_{j\in Y_{+}}log Pr(Y_{ij}=1|D^{Q}_{ij,m}) \\
                         &+\sum_{j\in Y_{-}}log Pr(Y_{ij}=0|D^{Q}_{ij,m}), 
\end{align}
where $Pr(Y_{ij}=1|D^Q_{ij,m})$ denotes that the prediction map in which each pixel denotes the affordance confidence. $Y_{+}$ and $Y_{-}$ denote the pixel sets of the affordance regions and non-affordance regions, respectively. The final training objective is defined as:
\begin{equation}
     \mathcal{L}=\sum_{i=1}^{N}\sum_{m=1}^{5}\mathcal{L}^{i,m}. \label{eq:no9}
\end{equation}

\section{The PADv2 Dataset}
\label{datasetsection}
In this section, we describe the details of the proposed  \textbf{P}urpose-driven \textbf{A}ffordance \textbf{D}ataset v2 (PADv2), including the process of collecting images, annotations and statistical analysis of the dataset. Some examples from our PADv2 dataset are shown in Fig. \ref{dataset_ins}, and
affordance masks and depth information annotations are shown in Fig. \ref{dataset example}.

\begin{figure*}[t]
	\centering
    \begin{overpic}[width=1.\linewidth]{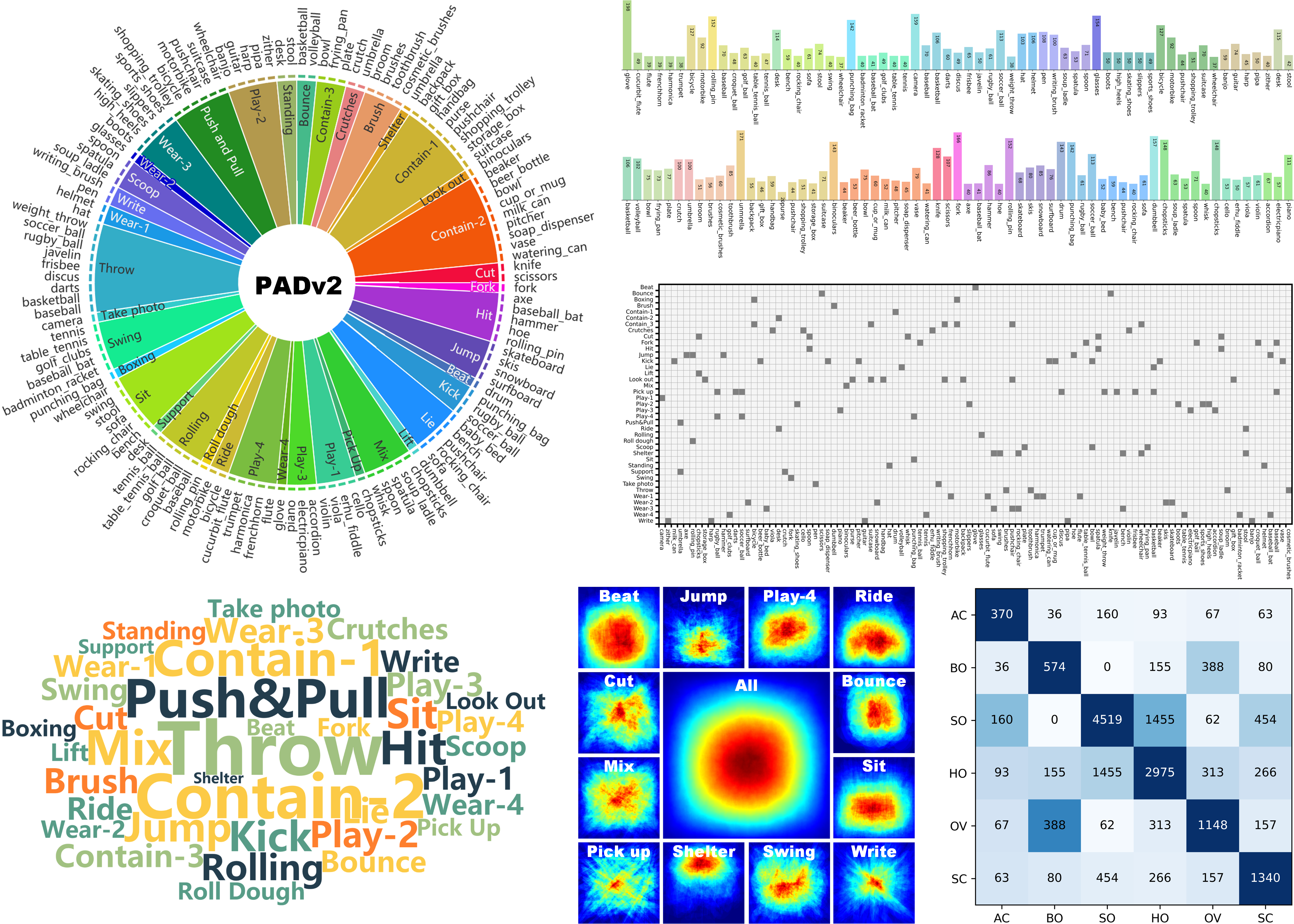}
        \put(0,30){\colorbox{black}{{\color{white} \footnotesize\textbf{(a)}}}}
        \put(0,0.5){\colorbox{black}{{\color{white} \footnotesize\textbf{(b)}}}}
        \put(46.5,51){\colorbox{black}{{\color{white} \footnotesize\textbf{(c)}}}}
        \put(46.5,28){\colorbox{black}{{\color{white} \footnotesize\textbf{(d)}}}}
        \put(43,0.5){\colorbox{black}{{\color{white} \footnotesize\textbf{(e)}}}}
        \put(73,0.5){\colorbox{black}{{\color{white} \footnotesize\textbf{(f)}}}}
    \end{overpic}
	\caption{\textbf{The properties of our dataset.} (a) The distribution of categories in the PADv2 dataset, which includes $39$ affordance categories and $103$ object categories. (b) The word cloud distribution of affordances in the PADv2 dataset. (c) Statistics of the number of images in each of the $103$ object classes in the PADv2 dataset. (d) Confusion matrix between the affordance category and the object category in the PADv2 dataset, where the horizontal axis denotes the object category and the vertical axis denotes the affordance category. (e) Visualization of the spatial distribution of afforance masks from specific affordance categories and the average mask of PADv2 dataset. (f) The distribution of co-occurring attributes of the PADv2 dataset. The number in each grid denotes the total number of images having a pair of specific attributes, which are described in detail as shown in Table \ref{attribute description}.}
	\label{dataset_abstract}
\end{figure*}

\begin{table*}[!t]
  \begin{center}
  \scriptsize
  \renewcommand{\arraystretch}{1.}
  \renewcommand{\tabcolsep}{6pt}
  \caption{Definition of the affordance categories in the PADv2 dataset and the object categories contained in each affordance category.}
  \label{Affordance Description table}
  \begin{tabular}{c|c|c}
    \toprule[1.2pt] 
  \textbf{Class} & \textbf{Description} & \textbf{Object Class} \\
   \midrule[1.2pt] 
   \rowcolor{mygray}
    Play-1     &    \multicolumn{1}{m{8.1cm}|}{Objects that make a sound when played by a person through a string as a medium.}     & \multicolumn{1}{m{6.1cm}}{violin, cello, erhu fiddle, viola}  \\\hline
    \rowcolor{mygray1}
    Play-2     &    \multicolumn{1}{m{8.1cm}|}{Objects that can interact directly with the hand to produce sound.}         & \multicolumn{1}{m{6.1cm}}{guitar, banjo, harp, pipa}  \\\hline
    \rowcolor{mygray}
    Play-3     &    \multicolumn{1}{m{8.1cm}|}{Objects that can produce sound by a person playing a keyboard.}      &  \multicolumn{1}{m{6.1cm}}{piano, accordion, electricpiano}          \\\hline
    \rowcolor{mygray1}
    Play-4      &    \multicolumn{1}{m{8.1cm}|}{Objects that a person blows through their mouth to make a sound.}    &  \multicolumn{1}{m{6.1cm}}{flute, trumpet, frenchhnorn, harmonica, cucurbit flute}   \\\hline
    \rowcolor{mygray}
    Take photo   &   \multicolumn{1}{m{8.1cm}|}{Objects that can take pictures of people.}      &  \multicolumn{1}{m{6.1cm}}{camera}           \\\hline
    \rowcolor{mygray1}  
    Contain-1   &   \multicolumn{1}{m{8.1cm}|}{Containers that can hold a variety of household items and miscellaneous goods.}        & \multicolumn{1}{m{6.1cm}}{backpack, gift box, handbag, purse, shopping trolley, suitcase, storage box, pushchair}   \\\hline
    \rowcolor{mygray}  
    Contain-2   &   \multicolumn{1}{m{8.1cm}|}{Objects that have the capable of containing a variety of liquids.}      & \multicolumn{1}{m{6.1cm}}{cup, bowl, beaker, pitcher, milk can, beer bottle, vase, soap dispenser, watering can} \\\hline
    \rowcolor{mygray1}  
    Contain-3  &   \multicolumn{1}{m{8.1cm}|}{Containers that can be used to hold food.}                      & \multicolumn{1}{m{6.1cm}}{bowl, frying pan, plate}      \\\hline
    \rowcolor{mygray}     
    Scoop     &   \multicolumn{1}{m{8.1cm}|}{Objects that have the specific ability to scoop food up from the bottom.}    &  \multicolumn{1}{m{6.1cm}}{spatula, spoon, soup ladle}        \\\hline
    \rowcolor{mygray1}   
    Wear-1   &    \multicolumn{1}{m{8.1cm}|}{Objects are worn on a person's head for decoration or protection.}            &  \multicolumn{1}{m{6.1cm}}{hat, helmet}            \\\hline
    \rowcolor{mygray}     
    Wear-2   &    \multicolumn{1}{m{8.1cm}|}{Objects with a combination of lenses and frames used to improve vision, or protect the eyes.}      &  \multicolumn{1}{m{6.1cm}}{glasses}                   \\\hline
    \rowcolor{mygray1}   
    Wear-3 & \multicolumn{1}{m{8.1cm}|}{Objects are worn on the feet.} &  \multicolumn{1}{m{6.1cm}}{high heels, boots, slippers, sports shoes} \\\hline
    \rowcolor{mygray}      
    Wear-4  & \multicolumn{1}{m{8.1cm}|}{Objects are worn on the hand for decoration, warmth or protection.} & \multicolumn{1}{m{6.1cm}}{gloves} \\\hline
    \rowcolor{mygray1}    
    Sit     &    \multicolumn{1}{m{8.1cm}|}{Objects that can be used to sit.}                 & \multicolumn{1}{m{6.1cm}}{stool, sofa, bench, swing, wheelchair, rocking chair}     \\\hline
    \rowcolor{mygray}      
    Cut      &  \multicolumn{1}{m{8.1cm}|}{Objects that have the ability to cut other  Obj. }        & \multicolumn{1}{m{6.1cm}}{knife, scissors}        \\\hline
    \rowcolor{mygray1}   
    Pick up   &  \multicolumn{1}{m{8.1cm}|}{Objects that have the function of holding food up.}         & \multicolumn{1}{m{6.1cm}}{chopsticks}              \\\hline
    \rowcolor{mygray}    
    Brush   &  \multicolumn{1}{m{8.1cm}|}{Objects that can remove dirt or apply cosmetics.}       & \multicolumn{1}{m{6.1cm}}{toothbrush, broom, brushes, cosmetic brushes}              \\\hline
    \rowcolor{mygray1}    
    Ride   &  \multicolumn{1}{m{8.1cm}|}{Vehicles that can be used for riding.}               & \multicolumn{1}{m{6.1cm}}{bicycle, motorbike}      \\\hline
    \rowcolor{mygray}      
    Kick   &  \multicolumn{1}{m{8.1cm}|}{Objects that can be kicked in direct contact with the foot.}        & \multicolumn{1}{m{6.1cm}}{soccer ball, punching bag, rugby ball}   \\\hline
    \rowcolor{mygray1}    
    Hit    &  \multicolumn{1}{m{8.1cm}|}{Indicates tools that can be used to strike other objects.}     & \multicolumn{1}{m{6.1cm}}{axe, hammer,  baseball bat, hoe, rolling pin}              \\\hline
    \rowcolor{mygray}      
    Beat   &  \multicolumn{1}{m{8.1cm}|}{Objects that can be played by beating a surface to produce a sound.}    & \multicolumn{1}{m{6.1cm}}{drum}                         \\\hline
    \rowcolor{mygray1}    
    Jump       &  \multicolumn{1}{m{8.1cm}|}{Objects that allow rapid movement by allowing people to jump on surfaces.}       & \multicolumn{1}{m{6.1cm}}{snowboard, surfboard, skis, skateboard}    \\\hline
    \rowcolor{mygray}      
    Swing      &  \multicolumn{1}{m{8.1cm}|}{Objects that a person interacts with by swinging their arm.}  & \multicolumn{1}{m{6.1cm}}{baseball bat, tennis racket, table tennis bat}   \\\hline
    \rowcolor{mygray1}    
    Lie        &  \multicolumn{1}{m{8.1cm}|}{Objects with a large surface space that allow a person to lie down.}     & \multicolumn{1}{m{6.1cm}}{sofa, baby bed, bench, pushchair, rocking chair}        \\\hline
    \rowcolor{mygray}      
    Bounce     &  \multicolumn{1}{m{8.1cm}|}{Objects that can be slapped directly by a person's hand and bounced.}          & \multicolumn{1}{m{6.1cm}}{basketball, volleyball}                     \\\hline
    \rowcolor{mygray1}    
    Mix        &  \multicolumn{1}{m{8.1cm}|}{Objects that can be used for mixing.}                 & \multicolumn{1}{m{6.1cm}}{whisk, chopsticks, spoon, soup ladle, spatula}    \\\hline
    \rowcolor{mygray}      
    Look out      &  \multicolumn{1}{m{8.1cm}|}{Objects that can be used for seeing at a distance.}   & \multicolumn{1}{m{6.1cm}}{binoculars}     \\\hline
    \rowcolor{mygray1} 
    Fork        &  \multicolumn{1}{m{8.1cm}|}{Indicates tools used to fork up food.}                          & \multicolumn{1}{m{6.1cm}}{fork}                 \\\hline
    \rowcolor{mygray}     
    Shelter      &  \multicolumn{1}{m{8.1cm}|}{Objects that provide shade from the sun or rain.}     & \multicolumn{1}{m{6.1cm}}{umbrella}            \\\hline
    \rowcolor{mygray1}  
    Roll dough    &  \multicolumn{1}{m{8.1cm}|}{Objects that can be used to roll out dough.}          & \multicolumn{1}{m{6.1cm}}{rolling pin}      \\\hline
    \rowcolor{mygray}  
    Rolling    &  \multicolumn{1}{m{8.1cm}|}{Objects that can be made to roll by hitting them through an intermediate medium.}   & \multicolumn{1}{m{6.1cm}}{table tennis ball, croquet ball, tennis ball, golf ball, baseball} \\\hline
    \rowcolor{mygray1}    
    Lift       &  \multicolumn{1}{m{8.1cm}|}{Objects that are lifted and lowered for fitness purpose.}      & \multicolumn{1}{m{6.1cm}}{dumbbell}          \\\hline
    \rowcolor{mygray}      
    Throw      &  \multicolumn{1}{m{8.1cm}|}{Objects that a person uses to throw.}                & \multicolumn{1}{m{6.1cm}}{frisbee, bowling,dart, javelin, basketball, weight throw, baseball, rugby ball}    \\\hline
    \rowcolor{mygray1}    
    Boxing  & \multicolumn{1}{m{8.1cm}|}{Objects hit by boxing sports.} & \multicolumn{1}{m{6.1cm}}{punching bag}  \\\hline
    \rowcolor{mygray}      
    Push\&Pull & \multicolumn{1}{m{8.1cm}|}{Objects with wheels below that can be pushed or pulled.} &    \multicolumn{1}{m{6.1cm}}{wheelchairs, bicycle, motorbike, suitcase, pushchair}  \\\hline
    \rowcolor{mygray1}    
    Crutches & \multicolumn{1}{m{8.1cm}|}{Objects that can play the function of assisting to walk people or stand.}  & \multicolumn{1}{m{6.1cm}}{crutch, umbrella} \\\hline
    \rowcolor{mygray}     
    Standing & \multicolumn{1}{m{8.1cm}|}{Objects that can be stepped on to reach a higher position.} & \multicolumn{1}{m{6.1cm}}{stool, desk, bench, sofa}  \\\hline
    \rowcolor{mygray1}    
    Support & \multicolumn{1}{m{8.1cm}|}{Objects with a smooth surface that can hold various items.} & \multicolumn{1}{m{6.1cm}}{desk} \\\hline
    \rowcolor{mygray}      
    Write & \multicolumn{1}{m{8.1cm}|}{Objects that can be used for writing.} & \multicolumn{1}{m{6.1cm}}{pen, writing brush} \\
    \bottomrule[1.2pt]
    \end{tabular}
    \end{center}
\end{table*}

\begin{table*}[t]
\caption{\textbf{The attribute list and associated description of the affordance object image.} The choice of these attributes is inspired by \cite{perazzi2016benchmark}.}
  \begin{center}
 \small
 \renewcommand{\arraystretch}{1}
  \renewcommand{\tabcolsep}{15pt}
  \begin{tabular}{c|c}
    \toprule[1.2pt] 
  Attr.   &   Description  \\
   \midrule[1.2pt]
   AC  &  \textbf{Appearance Change.} Significant lighting changes appear in the object area of the image.     \\
   BO  &  \textbf{Big Objects.} This refers to the ratio of object area to image area greater than 0.5.  \\
   HO  &  \textbf{Heterogeneus Object.} Refers to Obj. that are composed of visually distinct or dissimilar parts.  \\
   OV & \textbf{Out-of-View.}  Object is partially clipped by the image boundaries.\\
   SC & \textbf{Shape Complexity.}  The object has complex boundaries such asthin parts and holes. \\
   SO & \textbf{Small Object.} Refers to the ratio of object area to image area less than 0.1.\\
    \bottomrule[1.2pt]
    \end{tabular}
    \end{center}
    \label{attribute description}
\end{table*}

\subsection{Dataset Collection and Annotation}

In this section, we describe the collection and annotation process of our PADv2 dataset, where the annotations include the annotations of query images (masks), the annotations of support images (object mask, bounding boxes of the person and object, and human pose), and the depth map of each image.

\textbf{Data Collection.} We construct the PADv2 dataset by collecting images mainly from ILSVRC \citep{russakovsky2015imagenet}, COCO \citep{lin2014microsoft}, \etc. We retrieve and collect the images based on the keywords of object categories. The images in these datasets are from different scenes and have different object appearances, making the affordance detection task more challenging. In addition, to increase the diversity of the dataset, we also collect some images from the Internet. Finally, our dataset contains $30$k images, covering $39$ affordance categories and $103$ object categories. The affordance and the object categories are shown in Fig. \ref{dataset_abstract} (a). The word cloud statistics of all affordances are shown in Fig. \ref{dataset_abstract} (b) and the statistics of objects in each affordance category are shown in Fig. \ref{dataset_abstract} (c). Compared to PAD dataset \citep{Ours}, the PADv2 dataset is richer in content and more complex in scenarios. Furthermore, the multiple possibilities of object affordance in our PADv2 dataset make it more challenging and realistic for real-world application scenarios.
    
\textbf{Category Annotation.} We build a hierarchy for the PADv2 dataset by selecting $103$ common categories (\eg, ``bicycle'', ``stool'', ``tennis racket'', ``basketball'', ``binoculars'', ``chopsticks'', ``sofa'', ``soccer ball'') and assigning affordance category labels to each of them. The description of each affordance category of our PADv2 dataset and the object categories it contains are shown in Table \ref{Affordance Description table}. An affordance category may cover multiple object categories, \eg, objects having the ``Swing'' affordance label contain ``tennis racket'', ``golf club'', ``baseball bat'', ``badminton racket'', \etc., and the appearances of these objects vary greatly. Furthermore, an object may have multiple affordances. For example, the objects belong to the ``chopsticks'' category have both ``Pick Up'' and ``Mix'' affordances. Fig. \ref{dataset_abstract} (d) shows the confusion matrix between the affordance category and the object category.

\textbf{Query Image Annotation.} For the images from COCO \citep{lin2014microsoft}, part of the dataset is labeled with masks. Since the objects having the same affordance may have not been labeled in the above dataset, for example, the cups and bowls are not labeled in the category of cups, we filter these images and label them manually. For the images downloaded from the Internet, we also manually label the objects with the defined affordance categories. Some affordance masks from the PADv2 dataset are shown in Fig. \ref{dataset example}.
    
\textbf{Support Image Annotation.} Most of the support images come from the Internet, and we annotate the human-object interaction within each support image, \ie, the bounding box of the human and the bounding box of the interacting object. To facilitate the comparison with the one-shot segmentation methods, we also provide mask labels for the objects in the support images. Furthermore, we extract the human pose using \cite{zhang2021towards}, which can provide more information for inferring human action purpose.

\textbf{Depth Information.} As depth information provides a wealth of spatial structure, layout information, and geometric cues, it can facilitate the research of affordance detection. Therefore, we also provide the depth map of each image in the PADv2 dataset. We use the depth estimation network \citep{Ranftl2020} to extract the depth map. Some examples of depth maps are shown in Fig. \ref{dataset example}.

\subsection{Dataset Features and Statistics}

To get deeper insights into the PADv2 dataset, we show its important features from the following aspects.

\textbf{Category Diversity of Object.} PADv2 dataset contains $39$ affordance categories and $103$ object categories, including possible human-object interaction in outdoor, kitchen, living room, and other scenes. Each affordance category may cover multiple object categories. All the affordance categories cover most general objects in human life, supporting the research towards comprehensive understanding of the real-world scenes.

\textbf{The Multiple Possibilities of Affordance.} PADv2 dataset reflects the multiple possibility property of affordance, \ie, the same object may have multiple affordances and different categories of objects may have the same affordance. The confusion matrix between the affordance category and the object category is shown in Fig. \ref{dataset_abstract} (d), from which we can see the pairwise relationship between object and affordance. In this sense, our dataset is of great research value in affordance perception, scene understanding, and robotics.

\textbf{Spatial distribution of affordance masks.} Fig. \ref{dataset_abstract} (e) shows the average affordance mask of specific affordance category and the average mask of all affordance categories in the PADv2 dataset. It can be observed that some categories have unique location shapes. For example, ``Sit'' is mainly distributed in the lower part of the image, while ``Shelter" is mainly in the upper part of the image and has a distinctive shape. In contrast, ``Mix'', ``Write'', and ``Pick Up'' show no clear shape or position bias. The average mask of all categories is centered in the image with a circular shape.

\textbf{Property Analysis.} The attribute information of the images in the PADv2 dataset facilitate future research of model performance evaluation regarding different types of parameters. Inspired by \cite{perazzi2016benchmark}, we define a set of attributes to represent specific situations faced in real scenarios. Table \ref{attribute description} summarizes the list of attributes and their descriptions, and Fig. \ref{dataset_abstract} (f) shows the distribution of image attributes in the dataset. Since the real-world scenes consist of materials with different visual properties, ``HO'' types occupy a large portion. There are also many ``SO'' samples, implying that there are quite a few smaller objects in the dataset and its challenge nature for affordance detection. It can also be seen from the figure that ``HO'' and ``SO'' have a strong dependency. In addition, the number of images with ``OV'' attributes also makes up a large portion of the dataset, which also poses a huge challenge for one-shot affordance detection.

\section{Experiments}
\label{experiments-section}
In this section, we present the experimental settings, results and analysis. Section \ref{51} provides details of our benchmark setting. Section \ref{52} describes $11$ state-of-the-art methods from five relevant areas for comparison. In Section \ref{53}, we introduce the implementation details of our model. Section \ref{54} shows the experimental results and analyses. Specifically, Section \ref{541} presents the results of different models on the PADv2 dataset and the analysis from multiple perspectives. Section \ref{542} presents the results of multiple models on the PAD dataset. In Section \ref{543}, we investigate the impact of different modules in the proposed OSAD-Net and the hyper-parameter settings on the performance.

\begin{table*}[h]
\caption{\textbf{The division details of the PADv2 dataset for $3$-fold evaluation.} The table shows the affordance categories in each test set and the remaining part is the training set.}
\small
  \label{Table:1}
  \begin{center}
   \renewcommand{\arraystretch}{1}
  \renewcommand{\tabcolsep}{10pt}
  \begin{tabular}{c|c}
    \toprule[1.2pt] 
  Fold$_i$   &   Affordance Classes in the Test Set  \\
   \midrule[1.2pt]
   $i=1$  &  Bounce, Boxing, Contain-1, Crutches, Kick, Lie, Play-1, Push\&Pull, Ride, Shelter, Sit, Throw, Wear-1     \\
   \hline
   $i=2$  &  Brush, Contain-2, Fork, Hit, Jump, Lift, Mix, Pick Up, Play-2, Roll Dough, Scoop, Swing, Wear-2  \\
   \hline
   $i=3$  &  Beat, Contain-3, Cut, Look Out, Play-3, Play-4, Rolling, Standing, Support, Take Photo, Wear-3, Wear-4   \\
    \bottomrule[1.2pt]
    \end{tabular}
    \end{center}
    \label{divide}
\end{table*}

\subsection{Benchmark Setting}
\label{51}
To comprehensively evaluate different methods, we choose five widely used metrics for the One-Shot Affordance Detection task, \ie, IoU \citep{long2015fully}, F-measure ($F_{\beta}$) \citep{arbelaez2010contour}, E-measure ($E_\phi$) \citep{18IJCAI-Emeasure} , Pearson's Correlation Coefficient (CC) \citep{le2007predicting}, and Mean Absolute Error (MAE) \citep{perazzi2012saliency}. The evaluation code is released at \href{https://github.com/lhc1224/OSAD_Net/tree/main/PyMetrics/}{https://github.com/lhc1224/OSAD\_Net/PyMetrics/}.

\begin{itemize}
    \item 
    \textbf{IoU} \citep{long2015fully}. It is a critical metric commonly used to measure the results of pixel-level predictions and is calculated as follows:
    \begin{equation}
        IoU=\frac{X\cap Y}{X \cup Y},       
    \end{equation}
    where $X$ and $Y$ represent the prediction result and ground truth, respectively. $\cap$ is the intersection operation and $\cup$ is the union operation.
    \item 
    \textbf{F-measure ($F_{\beta}$)} \citep{arbelaez2010contour}. It is a broadly used evaluation metric that takes into account both recall and precision:
    \begin{align}
        &Recall=\frac{TP}{TP+FN}, \\
        &Precision=\frac{TP}{TP+FP}, \\
        &F_{\beta}=\frac{(1+\beta^2) Precision \cdot Recall}{\beta^2 \cdot Precision + Recall} \\
        &=\frac{(1+\beta^2)TP}{(1+\beta^2)TP+\beta^2FN+FP},
    \end{align}
    where TP indicates True Positives, FP represents False Positives, and FN means False Negatives. $\beta$ is a hyper-parameter to balance the recall and precision. In this paper we set $\beta = 0.3$.
    \item
    \textbf{E-measure ($E_\phi$)} \citep{18IJCAI-Emeasure}. It jointly evaluates the difference between the prediction result and ground truth from a local and global perspective:
    \begin{equation}
        E_\phi=\frac{1}{N}\sum_{n=1}^N\phi(X(n)-Y(n)),     
    \end{equation}
    where $\phi$ refers to the enhanced alignment matrix and $n$ is the index of each pixel.

    \item 
    \textbf{Pearson's Correlation Coefficient (CC)} \citep{le2007predicting}. It is a statistical metric usually used to count the correlation and dependence between two variables. In this paper, we use it to evaluate the correlation between the predicted map and ground truth:
    \begin{equation}
        CC(X,Y)=\frac{\sigma(X,Y)}{\sigma(X)\times \sigma(Y)},  
    \end{equation}
    where $\sigma(X,Y)$ is the covariance of $X$ and $Y$, and $\sigma(X)$ and $\sigma(Y)$ are the variances of $X$ and $Y$, respectively. CC is symmetric and penalizes false positives and negatives equally.
    
    \item
    \textbf{Mean Absolute Error (MAE)} \citep{perazzi2012saliency}. It measures the average absolute distance between the normalized predicted map and the ground-truth. 
    \begin{equation}
           MAE=\frac{1}{N}\sum_{n=1}^N|X(n)-Y(n)|.    \label{eq:no6}
    \end{equation}
\end{itemize}

\par To evaluate different models comprehensively, we follow the $k$-fold evaluation protocol, where $k$ is $3$ in this paper. To this end, the dataset is divided into three parts with non-overlapped categories, where any two of them are used for training while the left part is used for testing. The affordance categories included in each fold are shown in Table \ref{divide}. The training set contains $26$ affordance categories per fold, while the test set contains $13$ affordance categories.

\subsection{Comparison Methods}
\label{52}
To demonstrate the superiority of our model, we select three segmentation models (UNet, PSPNet, and DeeplabV3+), three salient object detection models (CPD, BASNet, and CSNet), one co-saliency model (CoEGNet), two few-shot segmentation models (CANet, and PFENet) and two affordance detection models (RANet, and OSAD-Net$^{\lozenge}$) for comparison.

\begin{itemize}
    \item 
    \textbf{UNet} \citep{10.1007/978-3-319-24574-4_28}: \textbf{U-Net} contains a contracting path to capture the semantic context and a symmetric expanding path to enable fine-grained localization.
    
    \item
    \textbf{PSPNet} \citep{zhao2017pspnet}: \textbf{P}yramid \textbf{S}cene \textbf{P}arsing \textbf{Net}work leverages the pyramid pooling module to aggregate contextual information from different receptive fields, thus improving the ability for semantic segmentation tasks.
    
    \item
    \textbf{DeepLabV3+} \citep{deeplabv3plus2018}:
    \textbf{DeepLabV3+} encodes multiscale features at different receptive fields by introducing an atrous spatial pyramid pooling (ASPP) module. Furthermore, it extends Deeplabv3 \citep{chen2017rethinking} by adding a decoding module to refine the segmentation results, especially for object boundaries.
    
    \item
    \textbf{CPD} \citep{Wu_2019_CVPR}: \textbf{C}ascaded \textbf{P}artial \textbf{D}ecoder discards the use of shallow layer information and proposes a cascaded encoding-decoding structure. 

 \item
    \textbf{BASNet} \citep{Qin_2019_CVPR}: \textbf{B}oundary-\textbf{A}ware \textbf{S}alient object detection \textbf{Net}work obtains a coarse result from an encoder-decoder structure and then refines it using a residual refinement module. In addition, it also utilizes a new loss, \ie, a mixture of the cross-entropy loss, the structural similarity loss, and the IoU loss.
    
    \item
    \textbf{CSNet} \citep{GaoEccv20Sal100K}: \textbf{C}ross-\textbf{S}tages \textbf{Net}work uses a generalized OctConv that effectively exploits multi-scale features within and across levels while reducing feature redundancy through a novel dynamic weight decay scheme.

 \begin{table*}[t]
\caption{\textbf{The experimental results of $11$ models} (UNet \citep{10.1007/978-3-319-24574-4_28}, PSPNet \citep{zhao2017pspnet}, DeeplabV3+ (DLabV3+) \citep{deeplabv3plus2018}, CPD \citep{Wu_2019_CVPR}, BASNet \citep{Qin_2019_CVPR}, CSNet \citep{GaoEccv20Sal100K}, CoEGNet \citep{deng2021re}, CANet \citep{zhang2019canet}, PFENet \citep{tian2020prior}, RANet \citep{zhao2020object}, OSAD-Net$^{\lozenge}$ \citep{Ours}) on the PADv2 dataset in terms of five metrics (IoU~$(\uparrow)$ \citep{long2015fully}, $F_{\beta}$~$(\uparrow)$ \citep{arbelaez2010contour}, $E_\phi$~$(\uparrow)$ \citep{18IJCAI-Emeasure}, CC~$(\uparrow)$ \citep{le2007predicting}, and MAE~$(\downarrow)$ \citep{perazzi2012saliency}). \textbf{Bold} and \uline{underline} indicate the best and the second-best scores, respectively.}
  \label{Table:2}
\centering
\small
  \renewcommand{\arraystretch}{1}
  \renewcommand{\tabcolsep}{2.pt}
  \begin{tabular}{c|r|ccc|ccc|c|cc|cc|c}
   \hline\toprule[1.2pt]
         & Method  & UNet & PSPNet & DLabV3+ & CPD  & BASNet & CSNet & CoEGNet & CANet & PFENet & RANet  & OSAD-Net$^{\lozenge}$  & Ours \\
         \hline
       \rowcolor{mygray1}  & Params (M) & $16.48$ & $50.85$  & $38.48$ & $45.63$ & $84.15$ & $0.752$ & $106.52$ & $18.15$ & $32.85$  & $46.18$ & $60.90$ & $42.77$ \\
     \midrule[1.2pt]
     \multirow{4}{*}{\rotatebox{90}{$i=1$}}  & IoU $(\uparrow)$  & $0.290$  & $0.279$  & $0.195$ & $0.381$ & $0.362$ & $0.285$ & $0.374$ & $0.402$ & $0.371$ & $0.301$  & $0.444$ & $0.496$ \\ 
      &  $F_{\beta}$ $(\uparrow)$ & $0.377$ & $0.411$ & $0.300$ & $0.488$ & $0.458$ & $0.390$ & $0.480$ & $0.417$ & $0.469$ & $0.393$ & $0.521$ & $0.565$  \\  
      & $E_{\phi}$ $(\uparrow)$ & $0.611$ & $0.631$ & $0.456$  & $0.675$ & $0.652$ & $0.600$ & $0.675$ & $0.600$ & $0.666$ & $0.598$ & $0.711$ & $0.745$ \\
      &    CC $(\uparrow)$   & $0.438$ & $0.476$ & $0.374$ & $0.526$ & $0.432$ & $0.506$ & $0.510$ & $0.512$ & $0.553$ & $0.502$ & $0.570$  & $0.611$   \\
      &   MAE $(\downarrow)$  & $0.163$ & $0.130$ & $0.126$ & $0.107$ & $0.112$ & $0.156$ & $0.107$ & $0.192$  & $0.106$ & $0.139$ & $0.115$ & $0.105$\\
      \hline
       \multirow{4}{*}{\rotatebox{90}{$i=2$}}  & IoU $(\uparrow)$ & $0.260$ & $0.278$ & $0.271$ & $0.357$ & $0.354$ & $0.282$ & $0.358$ & $0.347$ & $0.397$ & $0.306$ & $0.394$ & $0.432$  \\ 
      &  $F_{\beta}$ $(\uparrow)$ & $0.334$ & $0.370$ & $0.372$ & $0.437$ & $0.436$ & $0.351$ & $0.442$ & $0.392$ & $0.461$ & $0.366$ & $0.455$ & $0.492$ \\  
      & $E_{\phi}$ $(\uparrow)$ & $0.585$ & $0.609$ & $0.577$ & $0.656$ & $0.651$ & $0.578$ & $0.660$ & $0.623$ & $0.656$ & $0.579$ & $0.666$ & $0.689$ \\
      &    CC $(\uparrow)$  & $0.382$ & $0.417$ & $0.391$ & $0.465$ & $0.421$ & $0.435$ & $0.469$  & $0.458$  & $0.530$ & $0.437$ & $0.502$ & $0.542$ \\
      &   MAE $(\downarrow)$  & $0.185$ & $0.170$ & $0.139$ & $0.148$ & $0.141$ & $0.188$ & $0.140$ &  $0.164$ & $0.145$  & $0.195$ & $0.137$ & $0.140$  \\
      \hline
       \multirow{4}{*}{\rotatebox{90}{$i=3$}} & IoU $(\uparrow)$  & $0.347$ & $0.350$ & $0.292$ & $0.461$ & $0.440$ & $0.360$ & $0.471$ & $0.474$ & $0.481$ & $0.394$ & $0.490$ & $0.520$\\ 
      & $F_{\beta}$ $(\uparrow)$ & $0.450$ & $0.466$ & $0.428$ & $0.559$ & $0.536$ & $0.457$ & $0.575$ & $0.526$ & $0.565$ & $0.479$ & $0.563$  & $0.608$\\  
      & $E_{\phi}$ ($\uparrow$) & $0.624$ & $0.635$ & $0.571$ & $0.698$ & $0.686$ & $0.611$ & $0.716$ & $0.661$ & $0.697$  & $0.632$ & $0.695$ & $0.735$ \\
      & CC $(\uparrow)$ & $0.467$ & $0.509$ & $0.432$ & $0.560$ & $0.503$ & $0.538$ & $0.559$ & $0.600$  & $0.626$ & $0.545$ & $0.593$  & $0.609$ \\
      & MAE $(\downarrow)$ & $0.184$  & $0.163$ & $0.164$ & $0.139$ & $0.150$ & $0.181$ & $0.137$ & $0.160$ & $0.130$ & $0.169$ & $0.140$ & $0.125$ \\
      \hline
      \multirow{4}{*}{\rotatebox{90}{$Mean$}}  & IoU $(\uparrow)$ & $0.299$ & $0.302$ & $0.253$ & $0.400$ & $0.385$ & $0.309$ & $0.401$ & $0.408$ & $0.408$ & $0.334$ & $\uline{0.443}$ & $\textbf{0.483}$\\ 
      & $F_{\beta}$ $(\uparrow)$ & $0.387$ & $0.416$ & $0.358$ & $0.495$ & $0.477$ & $0.399$ & $0.499$ & $0.445$ & $0.494$ & $0.413$ & $\uline{0.513}$ & $\textbf{0.555}$\\ 
      & $E_{\phi}$ ($\uparrow$) & $0.607$ & $0.625$ & $0.535$ & $0.676$ & $0.663$ & $0.596$ & $0.684$ & $0.628$ & $0.673$ & $0.603$ & $\uline{0.691}$  & $\textbf{0.723}$  \\
      &    CC $(\uparrow)$ & $0.429$ & $0.467$ & $0.399$ & $0.517$ & $0.452$ & $0.493$ & $0.513$ & $0.523$ &  $\uline{0.570}$ & $0.495$  & $0.555$ & $\textbf{0.587}$ \\
      &   MAE $(\downarrow)$ & $0.177$ & $0.154$ & $0.143$ & $0.131$ & $0.134$ & $0.175$ & $0.128$  & $0.172$ & $\uline{0.127}$ & $0.168$ & $0.131$ & $\textbf{0.123}$ \\
      
    \bottomrule[1.2pt]
    \end{tabular}
    \label{result:PADv2}
  \end{table*}

    \item
    \textbf{CoEGNet} \citep{deng2021re}: \textbf{Co}- \textbf{E}dge \textbf{G}uidance \textbf{Net}work uses a co-saliency mapping strategy, \ie, the PCA technique, to identify the main components of common objects, helping to retain common objects and remove interference, thus enhancing the performance of EGNet \citep{zhao2019egnet} for co-salient object detection. 
    
    \item
    \textbf{CANet} \citep{zhang2019canet}: \textbf{C}lass-\textbf{A}gnostic segmentation \textbf{Net}work mainly contains a two-branch dense comparison module to compare the multi-scale features of support image and query image, and an iterative optimization module to refine the prediction results.

    \item
    \textbf{PFENet} \citep{tian2020prior}: \textbf{P}rior guided \textbf{F}eature \textbf{E}nrichment \textbf{Net}work uses a training-free prior generation to improve the segmentation accuracy and generalization performance, and a feature enrichment module to address spatial inconsistency.

    \item
    \textbf{RANet} \citep{zhao2020object}:  \textbf{R}elationship-\textbf{A}ware \textbf{Net}work improves the prediction results of affordance detection by exploring the relationship between affordance and objectness.
    
    \item
    \textbf{OSAD-Net$^{\lozenge}$} \citep{Ours}: \textbf{O}ne-\textbf{S}hot \textbf{A}ffordance \textbf{D}etection \textbf{Net}work$^{\lozenge}$ first learns the human action purpose, then transfers it to query images to obtain object regions with the same affordnace, and finally obtains results by collaborative learning.
    
\end{itemize}

\subsection{Implementation Details}
\label{53}
Our method is implemented in PyTorch. We choose the Resnet50 \citep{he2016deep} network pre-trained on ImageNet \citep{russakovsky2015imagenet} as the backbone, where the first three blocks are fixed. We randomly clip the input images from $360\times360$ to $320\times320$ with random horizontal flipping. We train the model for $35$ epochs using the Adam optimizer \citep{kingma2014adam}. The learning rate is initialized as $1e-4$ and reduced by $1/2$ after $15$ epochs. It takes about one day for training on a single NVIDIA TitanXP GPU. We set the batch size to $2$ and set the number of query images in each batch to $5$. For the action purpose learning module, we choose a four-layer GCN to process the skeletal data. The number of bases in the MPT module is $K$=$64$. The number of EM iteration steps is $5$.

\subsection{Analysis of Experimental Results}
\label{54}
In this section, we provide experimental results of all the comparison models on the PADv2/PAD datasets and analyze the experimental results. Then, we perform ablation studies to investigate the impact of different modules and hyper-parameter settings.

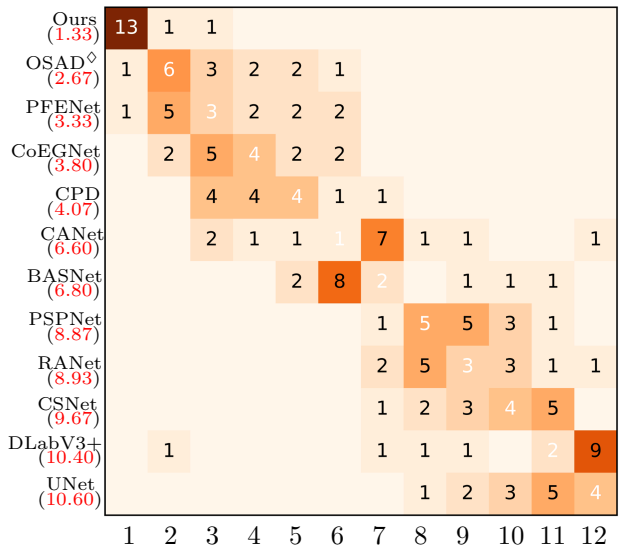
\begin{figure}[h]
    \begin{center}
      \begin{overpic}[width=0.92\linewidth]{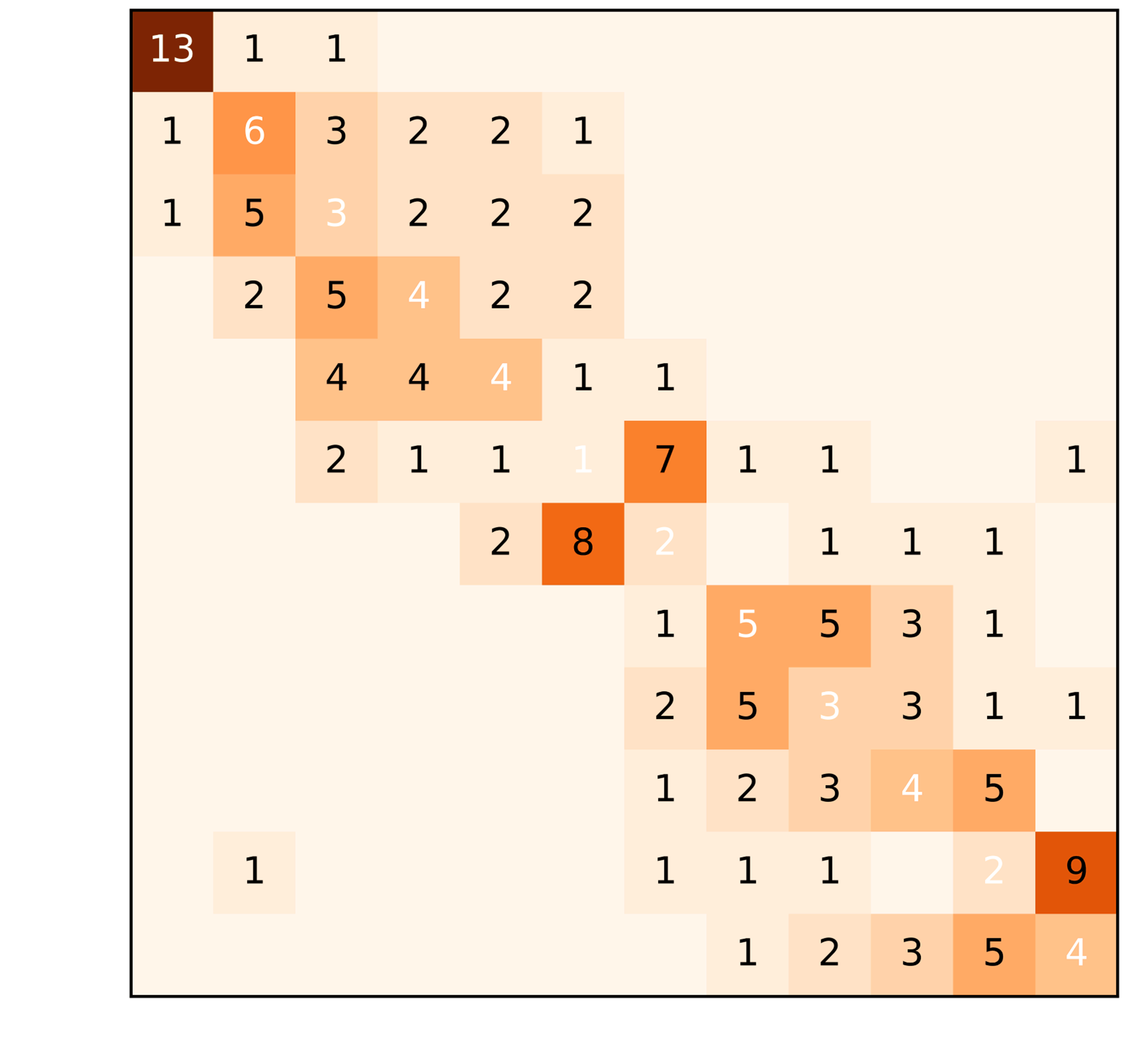}
           \put(2.5,90){\scriptsize Ours}
           \put(1.5,87.5){\scriptsize (\textcolor{red}{1.33})}
           
           \put(-2,82.4){\scriptsize OSAD$^{\lozenge}$}
           \put(1.5,79.9){\scriptsize (\textcolor{red}{2.67})}
           
           \put(-2.,75.){\scriptsize PFENet}
           \put(1.5,72.5){\scriptsize (\textcolor{red}{3.33})}
           
           \put(-4.5,67.4){\scriptsize CoEGNet}
           \put(1.5,64.9){\scriptsize (\textcolor{red}{3.80})}
           
           \put(3.,60){\scriptsize CPD}
           \put(1.5,57.5){\scriptsize (\textcolor{red}{4.07})}
           
           \put(0.5,53.2){\scriptsize CANet}
           \put(1.5,50.7){\scriptsize (\textcolor{red}{6.60})}
           
           \put(-1.8,46.){\scriptsize BASNet}
           \put(1.5,43.5){\scriptsize (\textcolor{red}{6.80})}
           
           \put(-1.5,38.5){\scriptsize PSPNet}
           \put(1.5,36){\scriptsize (\textcolor{red}{8.87})}
           
           \put(0.,31){\scriptsize RANet}
           \put(1.5,28.5){\scriptsize (\textcolor{red}{8.93})}
           
           \put(0.,24){\scriptsize CSNet}
           \put(1.5,21.5){\scriptsize (\textcolor{red}{9.67})}
           
           \put(-5,17.3){\scriptsize DLabV3+}
           \put(0.,14.8){\scriptsize (\textcolor{red}{10.40})}
           
           \put(1.5,10.2){\scriptsize UNet}
           \put(0.,7.7){\scriptsize (\textcolor{red}{10.60})}
           
           \put(92.5,1){12}
           \put(85.5,1){11}
           \put(78.4,1){10}
           \put(71.3,1){9}
           \put(64.2,1){8}
           \put(57.1,1){7}
           \put(50.0,1){6}
           \put(42.9,1){5}
           \put(35.8,1){4}
           \put(28.7,1){3}
           \put(21.6,1){2}
           \put(14.5,1){1}
        \end{overpic}
    \end{center}
    \caption{\textbf{Rank List.} We ranked the $15$ metrics of different methods in the $3$-fold evaluation on the PADv2 dataset (see Table \ref{result:PADv2} for experimental results), where the $element(i, j)$ denotes how many metrics that the model $i$ are ranked the $j$-th. The left \textcolor{red}{Red} letter denotes the average rank.}
    \label{rank image}
\end{figure}  

\begin{table*}[!t]
    \centering
    \renewcommand{\arraystretch}{1.}
    \renewcommand{\tabcolsep}{2.pt}
    \small
    \caption{\textbf{The results of the different methods on the PADv2 for each affordance category.} We use IoU as the evaluation metric. \textbf{Bold} and \uline{underline} indicate the best and the second-best scores, respectively.}
  \begin{tabular}{c|ccc|ccc|c|cc|c||c|c}
    \toprule[1.2pt] 
Classes  &   UNet &  PSPNet & DLabV3+  &  CPD &   BASNet  & CSNet  &  CoEGNet & CANet & PFENet & RANet & OSAD-Net$^{\lozenge}$  &  Ours  \\
   \midrule[1.2pt]
\rowcolor{mygray}
Play-1 & $0.367$ & $0.303$ & $0.361$ & $0.342$ & $0.331$ & $0.319$ & $0.375$ & $0.470$ & $0.473$ & $0.308$ & $\textbf{0.544}$ & $\uline{0.526}$ \\
\rowcolor{mygray1}
Play-2 & $0.378$ & $0.458$ & $0.429$ & $0.489$ & $0.443$ & $0.419$ & $0.530$ & $0.568$ & $\uline{0.620}$ & $0.475$ & $0.576$ & $\textbf{0.648}$ \\
\rowcolor{mygray}
Play-3 & $0.297$ & $0.276$ & $0.271$ & $0.387$ & $0.304$ & $0.334$ & $0.380$ & $0.414$ & $\uline{0.545}$ & $0.350$ & $0.443$ & $\textbf{0.552}$ \\   
\rowcolor{mygray1}
Play-4 & $0.356$ & $0.376$ & $0.409$ & $0.521$ & $0.456$ & $0.362$ & $0.547$ & $0.444$ & $0.501$ & $0.379$ & $\uline{0.572}$ & $\textbf{0.583}$ \\
\rowcolor{mygray}
Take Photo & $0.425$ & $0.37$ & $0.277$ & $0.488$ & $0.480$ & $0.438$ & $\uline{0.514}$ & $0.486$ & $0.446$ & $0.493$ & $0.464$ & $\textbf{0.517}$ \\
\rowcolor{mygray1}
Contain-1 & $0.294$ & $0.307$ & $0.137$ & $0.349$ & $0.352$ & $0.328$ & $0.322$ & $\uline{0.491}$ & $0.364$ & $0.329$ & $0.404$ & $\textbf{0.534}$ \\
\rowcolor{mygray}
Contain-2 & $0.384$ & $0.441$ & $0.391$ & $0.538$ & $0.492$ & $0.450$ & $0.538$ & $0.545$ & $\uline{0.564}$ & $0.490$ & $0.471$ & $\textbf{0.566}$ \\
\rowcolor{mygray1}
Contain-3 & $0.268$ & $0.323$ & $0.169$ & $0.325$ & $0.339$ & $0.285$ & $0.342$ & $\uline{0.526}$ & $0.482$ & $0.413$ & $\textbf{0.549}$ & $0.518$ \\
\rowcolor{mygray}
Scoop & $0.237$ & $0.242$  & $0.225$ & $0.297$ & $0.308$ & $0.261$ & $0.307$ & $0.305$ & $0.317$ & $0.271$ & $\textbf{0.381}$ & $\uline{0.372}$ \\
\rowcolor{mygray1}
Wear-1 & $0.340$ & $0.359$ & $0.243$ & $\uline{0.439}$ & $0.391$ & $0.390$ & $0.398$ & $0.435$ & $0.407$ & $0.390$ & $0.434$ & $\textbf{0.448}$ \\
\rowcolor{mygray}
Wear-2  &  $0.260$ & $0.230$ & $0.170$ & $\textbf{0.325}$ & $0.306$ & $0.270$ & $\uline{0.314}$ & $0.302$ & $0.257$ & $0.311$ & $0.262$ & $0.272$ \\
\rowcolor{mygray1}
Wear-3 & $0.469$ & $0.427$ & $0.274$ & $0.576$ & $0.552$ & $0.468$ & $0.522$ & $\textbf{0.630}$ & $0.548$ & $0.535$ & $0.557$ & $\uline{0.628}$ \\
\rowcolor{mygray}
Wear-4 & $0.249$ & $0.228$ & $0.100$ & $0.325$ & $\uline{0.354}$ & $0.254$ & $0.321$ & $0.449$ & $0.332$ & $0.293$ & $0.352$ & $\textbf{0.357}$ \\
\rowcolor{mygray1}
Sit & $0.251$ & $0.204$ & $0.101$ & $0.294$ & $0.309$ & $0.257$ & $0.303$ & $\uline{0.357}$ & $0.270$ & $0.223$ & $0.344$ & $\textbf{0.409}$ \\
\rowcolor{mygray}
Cut & $0.225$ & $0.245$ & $0.269$ & $0.363$ & $0.326$ & $0.221$ & $0.377$ & $0.334$ & $\uline{0.379}$ & $0.243$ & $\textbf{0.380}$ & $0.377$ \\
\rowcolor{mygray1}
Pick Up & $0.227$ & $0.205$ & $0.212$ & $0.289$ & $0.317$ & $0.240$ & $0.285$ & $0.259$ & $\textbf{0.442}$ & $0.224$ & $0.310$ & $\uline{0.314}$ \\
\rowcolor{mygray}
Brush     & $0.235$ & $0.256$ & $0.277$ & $0.354$ & $0.359$ & $0.263$ & $0.347$ & $0.319$ & $0.392$ & $0.235$ & $\uline{0.408}$ & $\textbf{0.429}$ \\
\rowcolor{mygray1}
Ride  & $0.314$ & $0.233$ & $0.066$ & $0.334$ & $0.316$ & $0.220$ & $0.308$ & $0.418$ & $0.374$ & $0.321$ & $\textbf{0.441}$ & $\uline{0.427}$ \\
\rowcolor{mygray}
Kick & $0.334$ & $0.406$ & $0.394$ & $0.546$ & $0.466$ & $0.330$ & $0.531$ & $0.493$ & $0.491$ & $0.413$ & $\uline{0.594}$ & $\textbf{0.665}$ \\
\rowcolor{mygray1}
Hit & $0.227$ & $0.258$ & $0.266$ & $0.361$ & $0.360$ & $0.216$ & $0.338$ & $0.319$ & $0.383$ & $0.242$ & $\uline{0.442}$ & $\textbf{0.486}$ \\
\rowcolor{mygray}
Beat  & $0.311$ & $0.416$ & $0.438$ & $0.433$ & $0.384$ & $0.307$ & $0.528$ & $0.579$ & $0.552$ & $0.432$ & $\textbf{0.650}$ & $\uline{0.643}$ \\
\rowcolor{mygray1}
Jump & $0.104$ & $0.085$ & $0.096$ & $0.190$ & $0.170$ & $0.083$ & $0.172$ & $0.162$ & $0.196$ & $0.165$ & $\uline{0.213}$ & $\textbf{0.272}$ \\
\rowcolor{mygray}
Swing  & $0.227$ & $0.187$ & $0.154$ & $0.252$ & $\uline{0.325}$ & $0.234$ & $0.275$ & $0.265$ & $0.292$ & $0.277$ & $\uline{0.325}$ & $\textbf{0.379}$ \\
\rowcolor{mygray1}
Lie & $0.323$ & $0.273$ & $0.128$ & $0.346$ & $0.383$ & $0.363$ & $0.357$ & $0.481$ & $0.391$ & $0.342$ & $\uline{0.433}$ & $\textbf{0.552}$ \\
\rowcolor{mygray}
Bounce & $0.238$ & $0.325$ & $0.232$ & $\uline{0.437}$ & $0.348$ & $0.238$ & $0.397$ & $0.280$ & $0.388$ & $0.280$ & $0.428$ & $\textbf{0.473}$ \\
\rowcolor{mygray1}
Mix & $0.237$ & $0.230$ & $0.222$  & $0.294$ & $0.317$ & $0.258$ & $0.296$ & $0.274$ & $\textbf{0.346}$ & $0.248$ & $\textbf{0.346}$ & $\uline{0.345}$ \\
\rowcolor{mygray}
Look Out & $0.443$ & $0.446$ & $0.481$ & $0.592$ & $0.544$ & $0.465$ & $\uline{0.600}$ & $0.508$ & $0.570$ & $0.517$ & $\textbf{0.603}$ & $\uline{0.600}$ \\
\rowcolor{mygray1}
Fork & $0.057$ & $0.064$ & $0.054$ & $0.078$ & $0.086$ & $0.061$ & $0.087$ & $0.063$ & $\uline{0.109}$ & $0.054$ & $\textbf{0.110}$ & $0.097$ \\
\rowcolor{mygray}
Shelter & $0.181$ & $0.103$ & $0.031$ & $0.176$ & $0.202$ & $0.208$ & $0.187$ & $0.270$ & $0.162$ & $0.102$ & $\uline{0.276}$ & $\textbf{0.355}$ \\
\rowcolor{mygray1}
Roll dough & $0.286$ & $0.334$ & $0.355$ & $0.392$ & $0.450$  & $0.300$ & $0.406$ & $0.398$ & $0.461$ & $0.307$ & $\uline{0.506}$ & $\textbf{0.553}$ \\
\rowcolor{mygray}
Rolling   & $0.377$ & $0.297$ & $0.193$ & $0.477$ & $\textbf{0.530}$ & $0.372$ & $\uline{0.479}$ & $0.371$ & $0.346$ & $0.402$ & $0.399$ & $0.427$ \\
\rowcolor{mygray1}
Lift  & $0.453$ & $0.491$ & $0.610$ & $0.603$ & $0.562$ & $0.490$ & $0.599$  & $0.535$ & $0.642$ & $0.532$ & $\uline{0.680}$ & $\textbf{0.691}$ \\
\rowcolor{mygray}
Throw & $0.223$ & $0.245$ & $0.210$ & $\uline{0.442}$ & $0.400$ & $0.203$ & $0.421$ & $0.281$ & $0.318$ & $0.269$ & $0.422$  & $\textbf{0.464}$\\
\rowcolor{mygray1}
Boxing & $0.456$  & $0.577$ & $0.629$ & $0.617$ & $0.499$ & $0.442$ & $0.662$ & $0.666$ & $0.716$ & $0.495$ & $\uline{0.731}$ & $\textbf{0.747}$ \\
\rowcolor{mygray}
Push and Pull  &  $0.329$ & $0.266$ & $0.088$ & $0.343$ & $0.350$ & $0.285$ & $0.321$ & $0.442$ & $0.366$ & $0.342$ & $\uline{0.461}$ & $\textbf{0.493}$ \\
\rowcolor{mygray1}
Crutches & $0.226$ & $0.117$ & $0.108$ & $0.245$ & $0.292$ & $0.246$ & $0.298$ & $0.259$ & $0.261$ & $0.146$ & $\uline{0.359}$ & $\textbf{0.377}$\\
\rowcolor{mygray1}
Standing & $0.343$ & $0.359$ & $0.239$ & $0.450$ & $0.449$ & $0.382$ & $0.478$ & $\uline{0.483}$ & $0.452$ & $0.330$ & $0.423$ & $\textbf{0.504}$ \\
\rowcolor{mygray}
Support & $0.373$ & $0.427$ & $0.240$ & $0.512$ & $0.491$ & $0.438$ & $\textbf{0.529}$ & $0.507$ & $\textbf{0.529}$ & $0.349$ & $0.420$ & $\uline{0.515}$ \\
\rowcolor{mygray1}
Write & $0.419$ & $0.468$ & $0.543$ & $0.598$ & $0.546$ & $0.433$ & $\uline{0.619}$ & $0.514$ & $\textbf{0.651}$ & $0.412$ & $0.613$ & $0.614$ \\
    \bottomrule[1.2pt]
    \end{tabular}
    \label{Table:IoU}
\end{table*} 

\begin{figure*}[!h]
	\centering
	\small
	\begin{overpic}[width=1\linewidth]{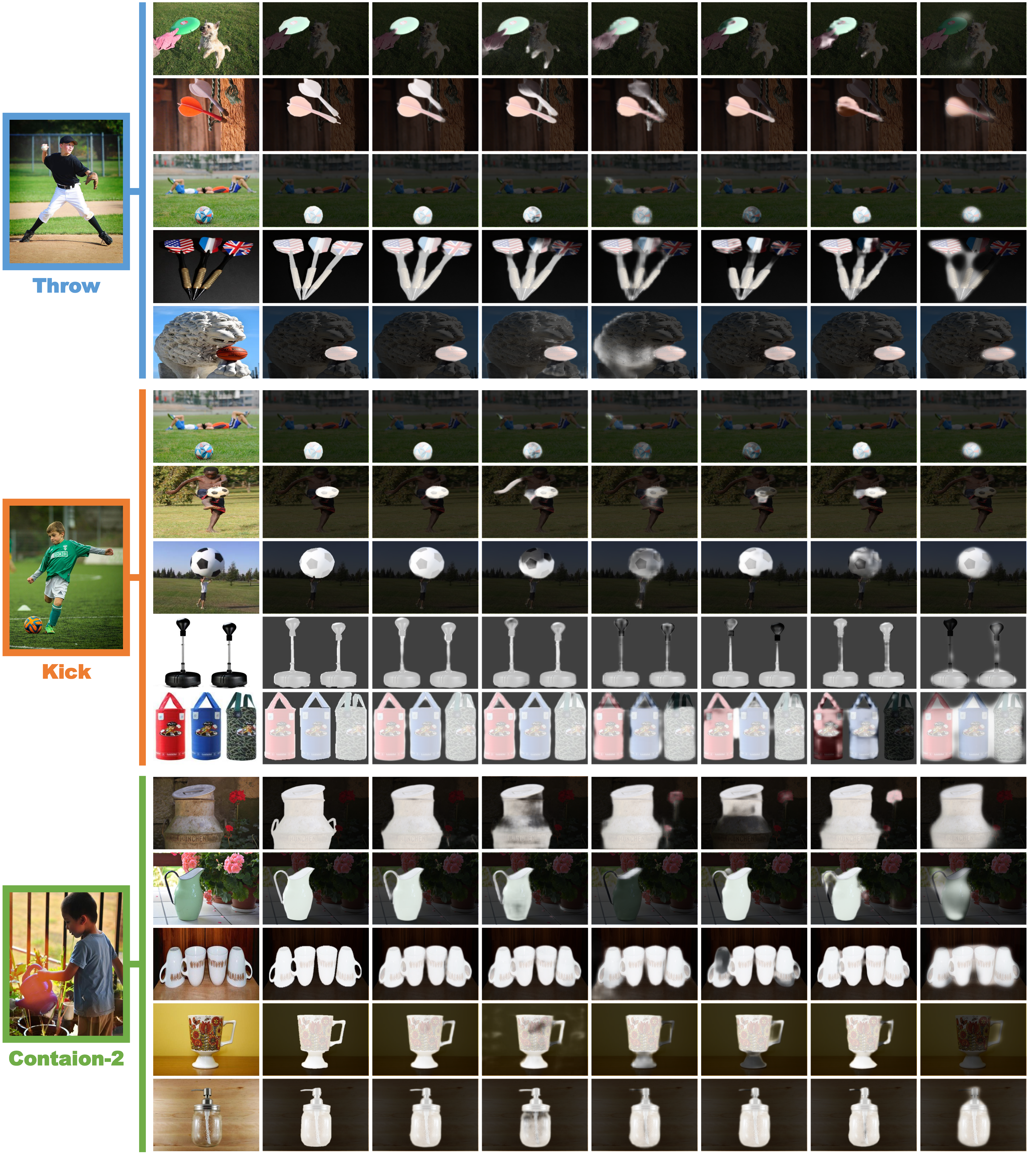}
		    \put(15.5, 0.2){\textbf{Image}}
		    \put(25.7, 0.2){\textbf{GT}}
		    \put(34.5, 0.2){\textbf{Ours}}
		    \put(41.3, 0.2){\textbf{OSAD-Net$^{\lozenge}$}}
		    \put(52.0, 0.2){\textbf{PFENet}}
		    \put(60.8, 0.2){\textbf{CoEGNet}}
		    \put(71.6, 0.2){\textbf{CPD}}
		    \put(80.0, 0.2){\textbf{PSPNet}}
	\end{overpic}
	\caption{\textbf{Visual affordance maps obtained by different methods on the PADv2.}}
	\label{FIG:MR}
\end{figure*}

\begin{figure*}[!h]
	\centering
		\begin{overpic}[width=1\linewidth]{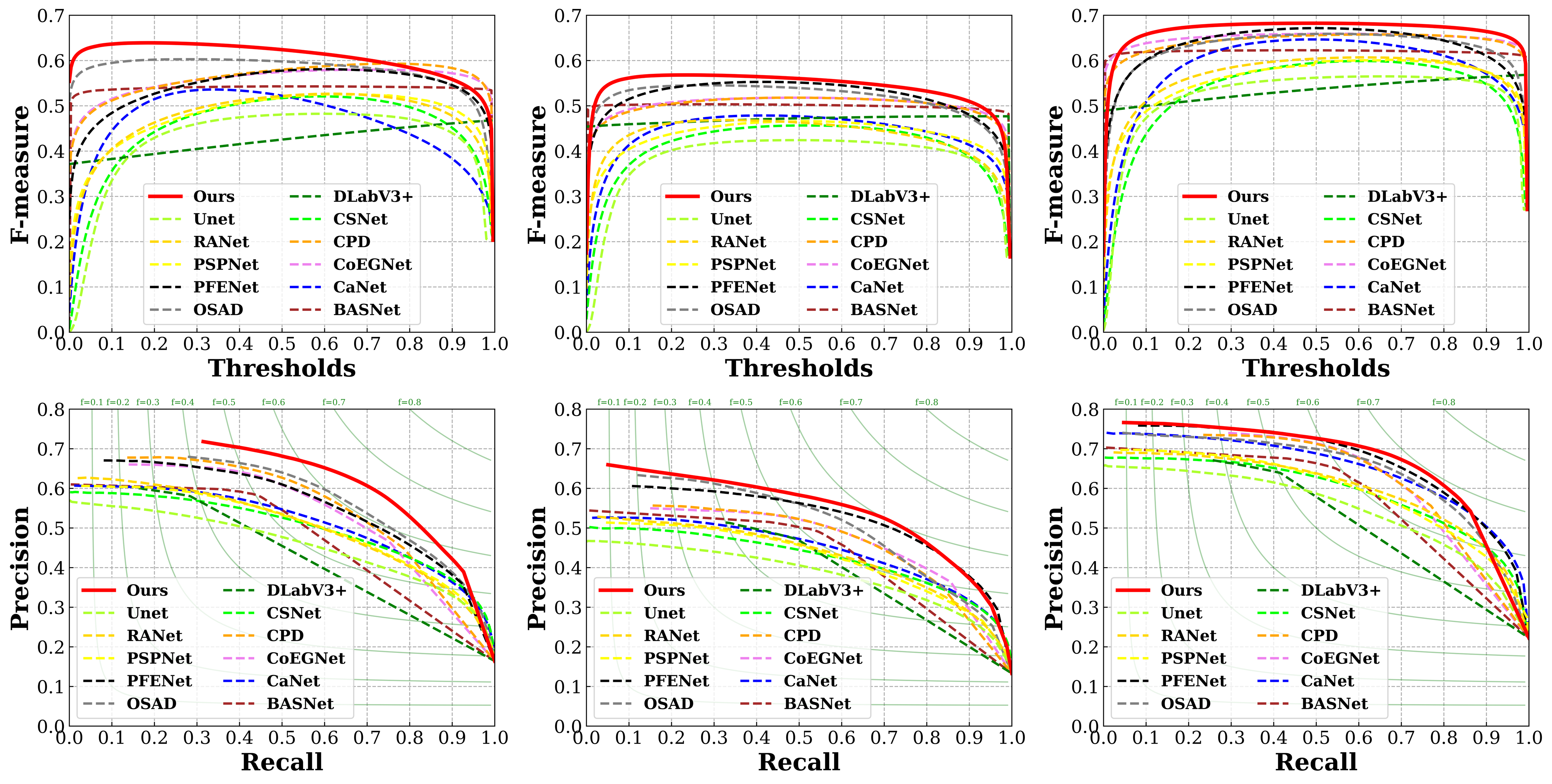}
		\put(4.2, 0.8){{\textcolor[RGB]{202,12,22}{\bm{$[i=1]$}}}}
		\put(4.2, 26.2){{\textcolor[RGB]{202,12,22}{\bm{$[i=1]$}}}}
		\put(37.6, 0.8){{\textcolor[RGB]{202,12,22}{\bm{$[i=2]$}}}}
		\put(37.6, 26.2){{\textcolor[RGB]{202,12,22}{\bm{$[i=2]$}}}}
		\put(70.9, 0.8){{\textcolor[RGB]{202,12,22}{\bm{$[i=3]$}}}}
		\put(70.9, 26.2){{\textcolor[RGB]{202,12,22}{\bm{$[i=3]$}}}}
		
		\put(11.4, 4.5){\scriptsize{$^{\lozenge}$}}
		\put(15.6, 29.9){\scriptsize{$^{\lozenge}$}}
		\put(44.7, 4.5){\scriptsize{$^{\lozenge}$}}
		\put(49.0, 29.9){\scriptsize{$^{\lozenge}$}}
		\put(78.1, 4.5){\scriptsize{$^{\lozenge}$}}
		\put(82.3, 29.9){\scriptsize{$^{\lozenge}$}}
		\end{overpic}
	\caption{\textbf{F-measure curves and PR curves of $12$ models on the PADv2 dataset.} The first and second rows are the F-measure and PR curves of the 3-fold evaluation ($[i=1,2,3]$), respectively.}
	\label{PR}
\end{figure*}

\subsubsection{Performance on the PADv2 dataset}
\label{541}
We compare our method with the representative methods of semantic segmentation, salient object detection, co-salient object detection, one-shot segmentation, and affordance detection as described in Section~\ref{52}. The experimental results are summarized in Table \ref{result:PADv2}. We also calculate the mean values of all metrics of the $3$-fold evaluation. As can be seen, our model outperforms all other methods in terms of all metrics. Taking the IoU metric as an example, our model outperforms the best affordance detection method by $9.03$\%, the best one-shot segmentation method by $18.38$\%, the co-salient object detection method by $20.45$\%, the best salient object detection method by $20.75$\%, and the best segmentation method by $59.93$\%. Furthermore, to compare the ranking of different methods in terms of all metrics, we rank all the evaluation metrics of the $3$-fold evaluation and summarize the result as a matrix as shown in Fig. \ref{rank image}, where the $(i, j)$ element indicates how many metrics the $i$-th model has ranked the $j$-th. Our method is most relevant to one-shot segmentation methods. The one-shot segmentation method requires the object mask of the support image, which provides finer grained information than the bounding boxes of the person and object but is hard to annotate. Nevertheless, our model still achieves a considerable improvement over the one-shot segmentation approach, indicating the effectiveness of our method that uses human action purpose to guide the network to predict affordance regions, rather than inferring affordance from object appearance only. 
\par We also show the IoU score of each method in each affordance category as shown in Table \ref{Table:IoU}, from which we can see that our model achieves the best scores on the affordance categories ``Contain-1'' and ``Contain-2'', which contain a wide range of objects with diverse appearances, and the best scores on the affordance categories ``Kick'' and ``Boxing'', which contains the same object category (\eg. ``punching bag''). These results confirm the ability of our model to address the challenges posed by the multiple possibilities of affordance. In addition, our model achieves the best results in most affordance categories, which shows that our model can adapt to a wide range of scenarios, \ie, it has a good generalization performance.

\par Meanwhile, we visualize the affordance maps predicted by each model in Fig. \ref{FIG:MR}. It is clear that our model can obtain more precise affordance masks for the same object belonging to different affordance categories (\eg, ``soccer ball'' belongs to both ``Throw'' and ``Kick'' affordance categories). It can also generate better segmentation masks on those affordance categories that cover multiple object categories with different appearances (\eg, ``frisbee'', ``darts'', ``basketball '', ``soccer ball'' all belong to ``Throw''). These results demonstrate that the proposed model can effectively learn the action purpose and mitigate the affordance ambiguity issue, \ie, multiple affordance possibilities collapse to an explicit affordance with the guidance of the transferred action purpose. Moreover, the mixture purpose transfer module and densely collaborative enhancement module in the proposed model also help to deal with the challenge of appearance differences between objects.

\par
\begin{figure}[t]
	\centering
		\begin{overpic}[width=1\linewidth]{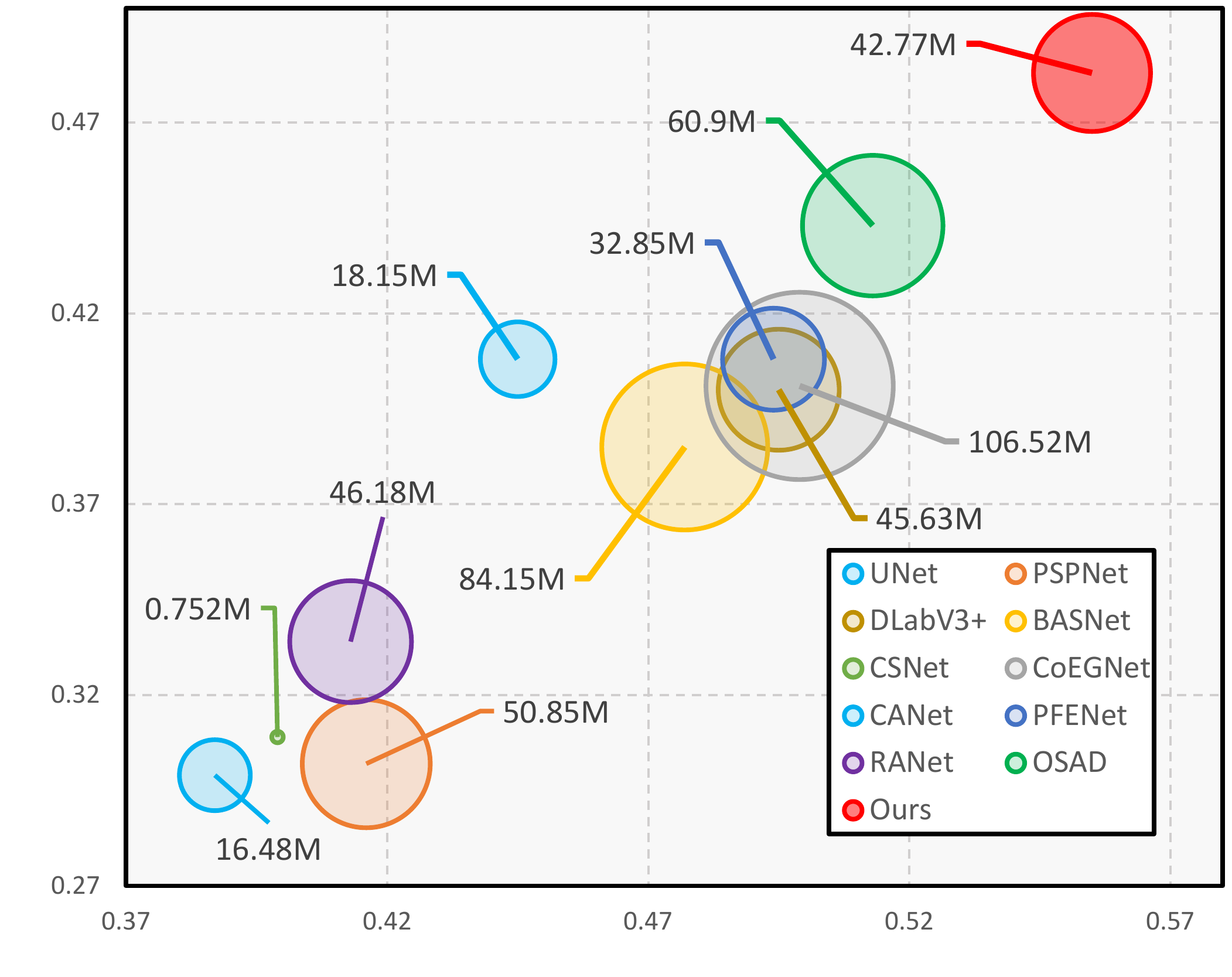}
		    \put(0.5, 40){\rotatebox{90}{\small\textbf{IoU}}}
		    \put(42.5, -0.2){\small\textbf{F-measure}}
		    \put(90, 16.2){\scriptsize{$^{\lozenge}$}}
	\end{overpic}
	\caption{\textbf{A dynamic bubble plot of model size (number of parameters) and model performance}, where the horizontal axis represents the F-measure ($F_\beta$) metric, the vertical axis represents the IoU metric, and the bubble size illustrates the number of parameters.}
	\label{IoUvsFmeasurevsmodels_size}
\end{figure}

 We plot PR curves of different methods by measuring the affordance maps using different thresholds varying from $0$ to $255$. For each threshold, a pair of recall and precision are calculated, from which we calculate the F-measure score. After varying the threshold, we can obtain the F-measure curve and the PR curve as shown in the first row and second row of Fig. \ref{PR}, respectively. As can be seen, the PR curve and F-measure curve of our model are at the top in most situations, demonstrating that it achieves the best performance on average. We also plot the dynamic bubble plots to analyze the relationship between the number of parameters and the model performance, as shown in Fig. \ref{IoUvsFmeasurevsmodels_size}, where the horizontal axis is the F-measure metric, and the vertical axis is the IoU metric. It shows that our model gets much higher IoU and F-measure scores than other models but with a smaller or comparable model size.

\subsubsection{Performance on the PAD dataset}
\label{542}
We also evaluate our model on the preliminary Purpose-driven Affordance Dataset (PAD) \citep{Ours}. The experimental results are shown in Table \ref{PAD result}, from which we can see that our model also achieves the best results in terms of several metrics of the $3$-fold evaluation. Although this dataset has a smaller scale, our model can also address the difficulties of affordance with multiple possibilities due to its better robustness. However, the improvement of our model over OSAD-Net$^{\lozenge}$ on the PAD dataset is not as significant compared to the PADv2 dataset. It is because that PADv2 dataset has more complex backgrounds, more diverse object categories, and more affordance categories, where the superiority of our model can be better evidenced.

\begin{table*}[t]
\centering
    \begin{minipage}[m]{0.24\textwidth}
    \centering
        \caption{\textbf{The experimental results of $7$ models} (UNet \citep{10.1007/978-3-319-24574-4_28}, PSPNet \citep{zhao2017pspnet}, CPD \citep{Wu_2019_CVPR}, BASNet \citep{Qin_2019_CVPR}, CSNet \citep{GaoEccv20Sal100K}, CoEGNet \citep{deng2021re}, OSAD-Net$^{\lozenge}$ \citep{Ours}) on the PAD dataset in terms of four metrics (IoU~$(\uparrow)$ \citep{long2015fully}, $E_\phi$~$(\uparrow)$ \citep{18IJCAI-Emeasure}, CC~$(\uparrow)$ \citep{le2007predicting}, and MAE~$(\downarrow)$ \citep{perazzi2012saliency}). \textbf{Bold} and \uline{underline} indicate the best and the second-best scores, respectively.}
         \label{PAD result}
    \end{minipage}
    \begin{minipage}[t]{0.75\textwidth}
    \centering
    \small
      \renewcommand{\arraystretch}{1.}
      \renewcommand{\tabcolsep}{3.5pt}
      \begin{tabular}{c|r|cc|ccc|c||c|c}
      \hline\toprule[1.2pt]
        & Metric  & UNet & PSPNet   & CPD   & BASNet  &  CSNet   & CoEGNet  & OSAD-Net$^{\lozenge}$ & Ours \\
        \hline
        \rowcolor{mygray1}  & Params (M) & $16.48$ & $50.85$  & $45.63$ & $84.15$ & $0.752$ & $106.52$ & $60.90$ & $42.77$ \\
    \midrule[1.2pt]
     \multirow{4}{*}{\rotatebox{90}{$i=1$}}  & IoU $(\uparrow)$  & $0.184$ & $0.231$ & $0.249$ & $0.239$  & $0.204$  & $0.277$  & $0.401$ & $0.424$ \\ 
          &    $E_{\phi}$ $(\uparrow)$ & $0.574$ & $0.640$ & $0.615$ & $0.604$ & $0.557$ & $0.674$ &  $0.732$ & $0.751$ \\  
          &    CC $(\uparrow)$ & $0.338$ & $0.427$ & $0.413$ & $0.310$  & $0.394$  & $0.389$ &    $0.540$ & $0.531$  \\
          &   MAE $(\downarrow)$ & $0.162$ & $0.144$ & $0.123$ & $0.130$ & $0.184$ &  $0.116$  &   $0.103$  & $0.098$ \\
    \hline
     \multirow{4}{*}{\rotatebox{90}{$i=2$}}  & IoU $(\uparrow)$ & $0.215$ & $0.227$ & $0.251$   &  $0.263$  &  $0.210$  & $0.259$ &  $0.375$ & $0.411$ \\ 
          &    $E_{\phi}$ $(\uparrow)$ & $0.558$  & $0.601$ &  $0.601$ &  $0.598$  & $0.555$  & $0.637$ &  $0.653$ & $0.684$  \\  
          &    CC $(\uparrow)$ & $0.377$  & $0.409$ & $0.386$ & $0.318$  &  $0.392$ & $0.350$ &  $0.507$ & $0.502$   \\
          &   MAE $(\downarrow)$  & $0.163$  & $0.142$ & $0.106$  & $0.124$ & $0.162$  &  $0.110$ &  $0.116$ & $0.109$   \\
    \hline
     \multirow{4}{*}{\rotatebox{90}{$i=3$}}  & IoU $(\uparrow)$  & $0.227$ & $0.265$  & $0.285$  & $0.281$ &  $0.238$   & $0.284$  &  $0.407$ & $0.410$  \\ 
          &    $E_{\phi}$ $(\uparrow)$ & $0.578$ & $0.636$  &  $0.630$  &  $0.628$ & $0.557$  & $0.645$  & $0.691$ & $0.665$   \\  
          &    CC $(\uparrow)$ & $0.344$ & $0.402$  &  $0.433$  & $0.339$  &  $0.386$  &  $0.362$  &  $0.501$ & $0.494$    \\
          &   MAE $(\downarrow)$ & $0.169$  & $0.137$  &  $0.132$  &  $0.146$  &  $0.184$  &   $0.134$  & $0.122$ & $0.130$  \\
    \hline
          \multirow{4}{*}{\rotatebox{90}{$Mean$}}  & IoU $(\uparrow)$  & $0.209$ & $0.241$ & $0.262$ & $0.261$ &  $0.217$ & $0.273$ &  $\uline{0.394}$ & $\textbf{0.415}$ \\
          &    $E_{\phi}$ $(\uparrow)$ & $0.570$ & $0.626$ & $0.615$ &  $0.610$ & $0.556$  & $0.652$ & $\uline{0.691}$ & $\textbf{0.700}$ \\  
          &    CC $(\uparrow)$  & $0.353$ & $0.413$ & $0.411$ & $0.322$ & $0.391$ & $0.367$  &  $\textbf{0.516}$ & $\uline{0.510}$   \\
          &   MAE $(\downarrow)$ & $0.165$ & $0.141$ & $0.120$ & $0.133$ & $0.177$ & $0.120$ &  $\uline{0.114}$ & $\textbf{0.112}$  \\
    \bottomrule[1.2pt]
    \end{tabular}
    \end{minipage}
\end{table*}

\begin{table*}[t]
    \begin{minipage}[m]{0.24\textwidth}
        \caption{\textbf{Ablation study results on the PADv2 dataset.} We investigate the impact of the APL module, the MPT module, and the DCE module in the proposed model.}
         \label{Table:ablation study}
    \end{minipage}
    \begin{minipage}[t]{0.75\textwidth}
    \centering
    \small
      \renewcommand{\arraystretch}{1.}
      \renewcommand{\tabcolsep}{6.5pt}
  \begin{tabular}{c|ccc|ccccc}
    \toprule[1.2pt]
  &  APL              &    MPT       &      DCE       &  IoU $(\uparrow)$ & $F_{\beta}$ $(\uparrow)$ & $E_{\phi}$ $(\uparrow)$ & CC ($\uparrow$) & MAE ($\downarrow$)       \\
    \midrule[1.2pt]
  (a) &                   &              &                &  $0.336$ & $0.417$   & $0.598$ & $0.498$ & $0.137$   \\
  (b) &  \cmark       &              &                &  $0.375$ & $0.454$ & $0.618$ & $0.553$ & $0.133$   \\
    (c) &                 &              &   \cmark   & $0.397$ & $0.494$ & $0.701$ & $0.493$ & $0.125$       \\
 (d) &   \cmark       &  \cmark  &                & $0.436$ & $0.491$ & $0.658$ & $0.583$ & $0.127$       \\
  (e) &  \cmark       &              &   \cmark   &    $0.451$ & $0.504$ & $0.685$ & $0.581$ & $0.124$    \\
    \midrule
    \rowcolor{mygray}
Ours &    \cmark       &  \cmark  &   \cmark   &  $\textbf{0.483}$ & $\textbf{0.555}$  & $\textbf{0.723}$  & $\textbf{0.587}$ & $\textbf{0.123}$     \\
    \bottomrule[1.2pt]
    \end{tabular}
    \end{minipage}
\end{table*}

\begin{figure*}[t]
	\centering
	\small
	\begin{overpic}[width=1\linewidth]{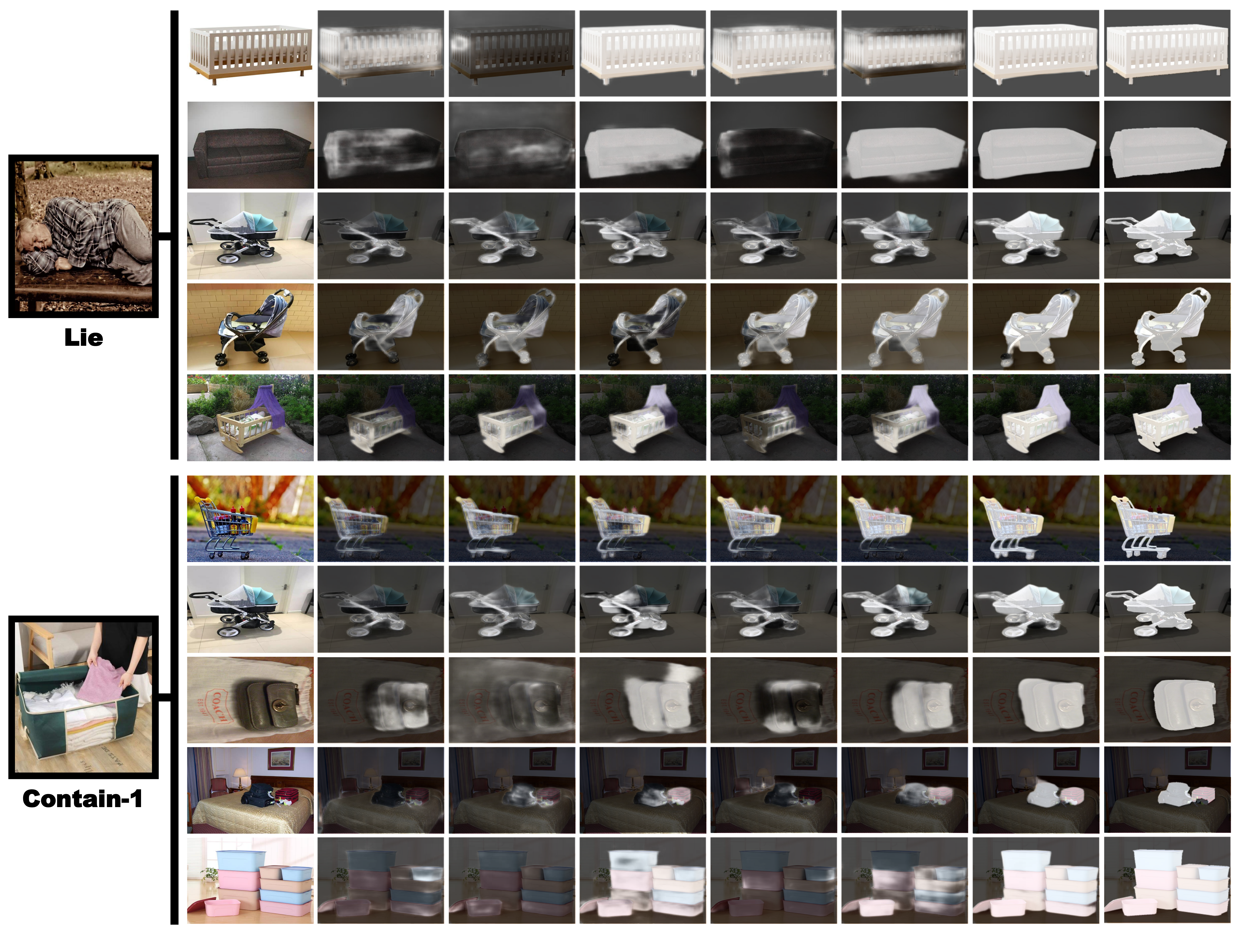}
		    \put(17.5, 0.2){\textbf{Image}}
		    \put(30.0, 0.2){\textbf{(a)}}
		    \put(40.2, 0.2){\textbf{(b)}}
		    \put(51.0, 0.2){\textbf{(c)}}
		    \put(61.5, 0.2){\textbf{(d)}}
		    \put(72.3, 0.2){\textbf{(e)}}
		    \put(82., 0.2){\textbf{Ours}}
		    \put(93.2, 0.2){\textbf{GT}}
	\end{overpic}
	\caption{\textbf{Visualization of affordance maps in the ablation studies on the PADv2 dataset.} (a) The network does not contain any of the proposed modules. (b) The network only uses the APL module. (c) The network only uses the DCE module. (d) The network uses the APL module and MPT module. (e) The network use the APL module and DCE module.}
	\label{ablation_result maps}
\end{figure*}

\subsubsection{Ablation Study}
\label{543}
In this section, we investigate the impact of each module in the proposed model on the one-shot affordance detection task. Further, we also study the influence of different hyper-parameter settings on the model performance, including the the number of input query images $N$, the number of optimization iterations, the number of bases in the MPT module, and the calculation of correlations between images in the DCE module.

\par To investigate the impact of the APL module, the MPT module, and the DCE module, we evaluate all combinations of them. Since the MPT is only necessary when the APL is present, there are five combinations in total. Note that in two sets of experiments, \ie, APL and APL+DCE, we resize the output of MPT directly to the scale of query image features after averaging pooling, concatenate it with query image features, and then pass them through a convolution layer to get the output. The experimental results are listed in Table \ref{Table:ablation study}. In addition, we also show some visual results in Fig. \ref{ablation_result maps}. From these results, it can be seen that the DCE module can bring a considerable improvement by learning the common relationship between objects of the same affordance, which can effectively mitigate the influence of the appearance differences between objects of the same affordance. From the visual results, we can see that the DCE module helps the network to obtain a more accurate segmentation mask with complete structure of the object. Besides, the APL module and the MPT module can also improve the results, since they can roughly activate the relevant regions of the object. However, the results are relatively coarse, which in turn shows that the DCE can help to obtain more accurate segmentation masks by effectively learning the common feature between objects having the same affordance.

\par
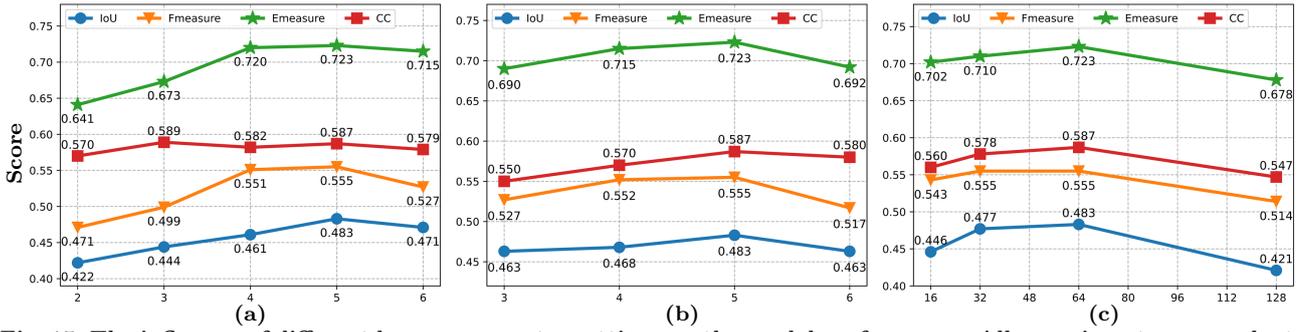
\begin{figure*}[!h]
	\centering
	\begin{overpic}[width=0.97\linewidth]{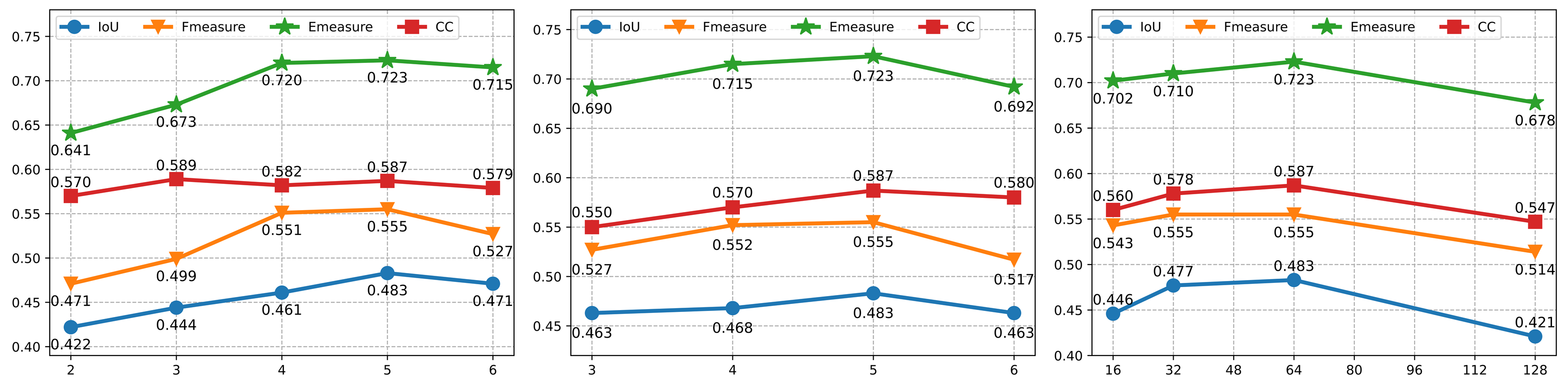}
		    \put(-1.0, 10){\rotatebox{90}{\footnotesize\textbf{Score}}}
		    \put(16.7, -0.7){\small\textbf{(a)}}
		    \put(50.3, -0.7){\small\textbf{(b)}}
		    \put(83.3, -0.7){\small\textbf{(c)}}
	\end{overpic}
	\caption{\textbf{The influence of different hyper-parameter settings on the model performance. All experiments are conducted on the PADv2 dataset.} (a) The influence of the number of query images. (b) The influence of the number of iterations of optimization in the MPT module . (c) The influence of the number of bases in the MPT module.}
	\label{Ablation study N}
\end{figure*}

\begin{figure}[!h]
	\centering
		\begin{overpic}[width=1\linewidth]{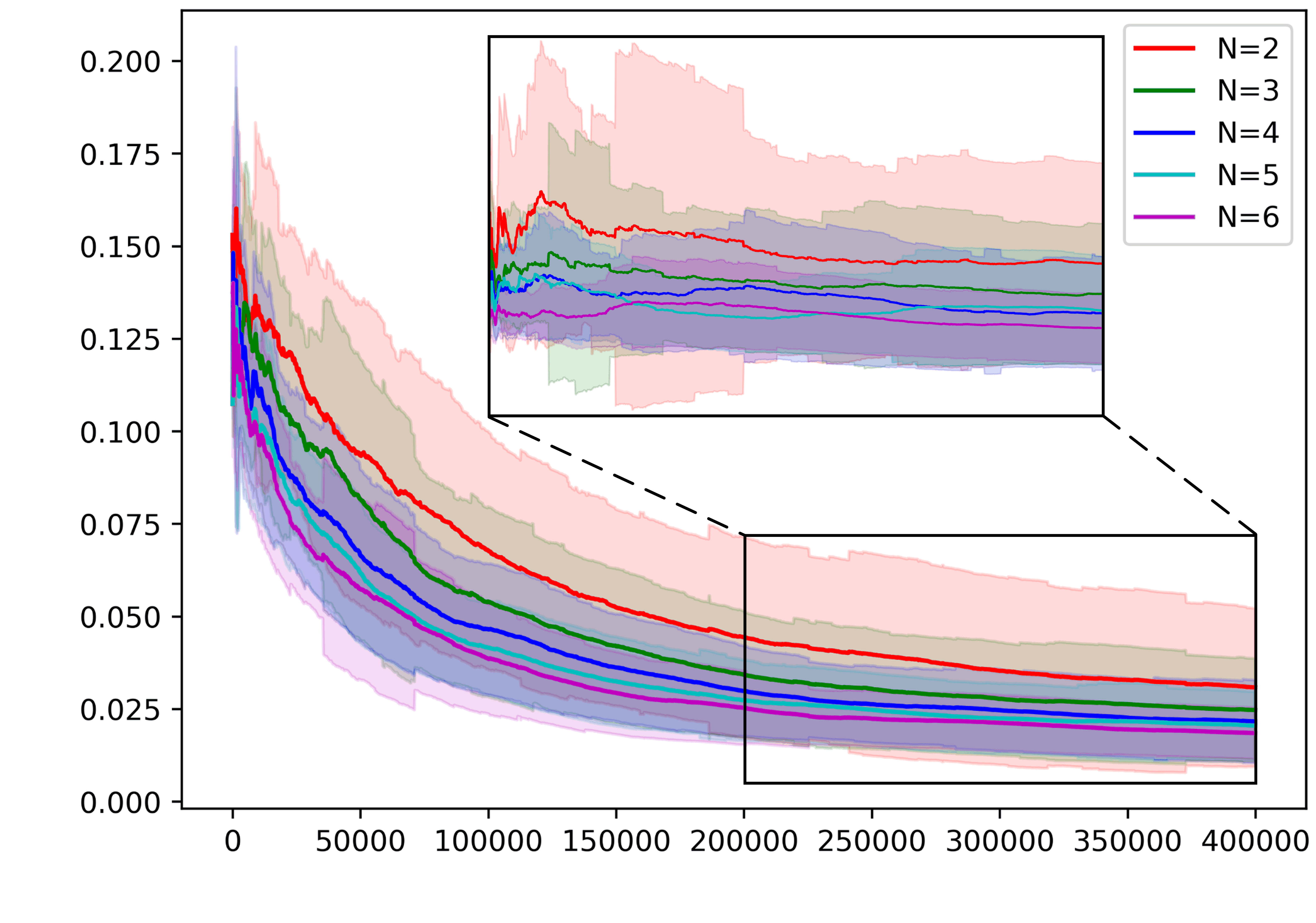}
		    \put(1, 35){\rotatebox{90}{\small\textbf{Loss}}}
		    \put(47, 0.5){\small\textbf{Iterations}}
	\end{overpic}
	\caption{\textbf{The training loss curves} for different numbers of query images, \ie, $N=2, 3, 4, 5, 6$. 
	}
	\label{Loss_N}
\end{figure}

 Fig. \ref{Ablation study N} (a)$\sim$(c) show the influence of the number of query images $N$, the number of optimization iterations of the Gaussian mixture model, and the number of Gaussian mixture module bases on the performance of the proposed model, respectively. From Fig. \ref{Ablation study N} (a), it suggests that the number of query images has a significant impact on the performance, \ie, all four metrics are improved as $N$ increases. Meanwhile, Table \ref{Table:ablation studyN} shows the scores at different settings of $N$ and OSAD-Net$^{\lozenge}$. As can be seen, our model with $N=3$ already surpasses OSAD-Net$^{\lozenge}$ in terms of several metrics, demonstrating that it has higher data efficiency and can exploit the common feature of query images more effectively than OSAD-Net$^{\lozenge}$. We also find when $N=6$ does not bring a significant improvement compared to $N=5$. It suggests that as $N$ increases, the effectiveness of the model may be reaching a bottleneck. Moreover, we also plot the training losses for different numbers of query images in Fig. \ref{Loss_N}, from which we have three observations: (1) the loss decreases as $N$ increases; (2) the stability of the loss increases obviously as $N$ increases; and (3) as $N$ increases, the decrease in loss becomes less pronounced. It shows that increasing the number of query images can effectively improve the stability and performance of network training. However, this improvement becomes insignificant when the number of $N$ keeps increasing. It may be because more query images introduce redundancy of information and an increase in computational complexity, resulting in no significant improvement in the network's performance.

Besides, the number of iterations of optimization in the MPT module does not significantly impact the performance, and the metrics improve slightly as the number of iterations increases. The best results are obtained when the number of iterations is $5$. The influence of the number of bases in the MPT module is shown in Fig. \ref{Ablation study N} (c). The best results peak at $K=64$, while the metrics decrease significantly when the number of bases increases to $128$, implying that using more bases may lead to an overfitting issue and the model becomes prone to noise. In Table \ref{Table:ablation study3}, we also compare the influence of different ways of calculating the similarity of two vectors in the DCE module. It shows that using cosine similarity is much better than Embedded Gaussian. It is because cosine similarity is not sensitive to absolute values and only measures the direction of the two vectors. Therefore, for regions of objects that are not significantly activated, the likelihood of them belonging to the same affordance object is still large by using the cosine similarity, making them also contribute to the learning process.
\begin{table}[!h]
  \renewcommand{\arraystretch}{1.}
  \caption{\textbf{The influence of the number of query images ($N$) in the DCE module on the performance.} 
  }
  \label{Table:ablation studyN}
  \begin{center}
  \centering
  \small
      \renewcommand{\arraystretch}{1.}
      \renewcommand{\tabcolsep}{1.pt}
  \begin{tabular}{c|ccccc}
    \toprule[1.2pt]
        $N$   &  IoU $(\uparrow)$ & $F_{\beta}$ $(\uparrow)$ & $E_{\phi}$ $(\uparrow)$ & CC ($\uparrow$) & MAE ($\downarrow$)       \\
    \midrule[1.2pt]
    OSAD-Net$^{\lozenge}$ $(N=5)$ & $0.443$ & $0.513$ & $0.691$ & $0.555$ & $0.131$ \\
    \hline
     $N=2$   & $0.422$ & $0.471$ & $0.641$ & $0.570$ & $0.141$       \\
     $N=3$     & $0.444$  & $0.499$ & $0.673$ & $\textbf{0.589}$ & $0.127$    \\
     $N=4$    &  $0.461$ & $0.551$ & $0.720$ & $0.582$ & $0.128$     \\
     \rowcolor{mygray}
    $N=5$ (Ours)     & $\textbf{0.483}$ & $\textbf{0.555}$ & $\textbf{0.723}$ & $0.587$ & $0.123$   \\
    $N=6$ &  $0.471$ & $0.527$ & $0.715$ & $0.579$ & \textbf{0.120}\\
    \bottomrule[1.2pt]
    \end{tabular}
    \end{center}
\end{table}

\section{Conclusion and Discussion}
\label{conclusionsection}
In this paper, we make the first attempt to address the challenging task named one-shot affordance detection, which is of great significance for real-world applications, such as empowering robots with the ability to perceive unseen affordance. Specifically, we propose a novel One-Shot Affordance Detection Network (OSAD-Net), which shows excellent ability at addressing the affordance ambiguity issue and recognizing unseen affordance effectively. With the proposed action purpose learning module, it can leverage the human action purpose to infer the object affordance in the environmental context and transfer it to the query images via a mixture purpose transfer module, and efficiently discover the common affordance feature of different objects in query images via a collaborative learning strategy. In addition, we also establish a large-scale challenging benchmark named Purpose-driven Affordance Dataset v2 (PADv2), which contains $30$k images covering $39$ affordance categories and $103$ object categories from complex real-world scenarios, and it can serve as a test bed to facilitate future research in this area. Comprehensive evaluation results confirm the superiority of the proposed OSAD-Net over representative models adapted from related areas, including semantic segmentation, salient object detection, co-salient object detection, one-shot segmentation, and affordance detection, making it a strong baseline for this challenging task.

\begin{table}[!t]
  \renewcommand{\arraystretch}{1.}
  \caption{\textbf{Comparison of different similarity calculation methods in the DCE module.}
  }
  \begin{center}
  \centering
  \small
      \renewcommand{\arraystretch}{1.}
      \renewcommand{\tabcolsep}{0.8pt}
  \begin{tabular}{c|ccccc}
    \toprule[1.2pt]
           &  $\text{IoU}$~$(\uparrow)$ & $F_{\beta}$~$(\uparrow)$ & $E_{\phi}$~$(\uparrow)$ & $\text{CC}$~$(\uparrow)$ & $\text{MAE}$~$(\downarrow)$       \\
    \midrule[1.2pt]
    Embedded Gaussian   & $0.442$ & $0.524$ & $0.702$  & $0.551$ & $\textbf{0.130}$      \\
    \rowcolor{mygray}
     Cosine similarity     & $\textbf{0.483}$ & $\textbf{0.555}$ & $\textbf{0.723}$ & $\textbf{0.587}$ & $0.123$     \\
    \bottomrule[1.2pt]
    \end{tabular}
    \end{center}
  \label{Table:ablation study3}
\end{table}

\begin{figure}[!t]
	\centering
		\begin{overpic}[width=1\linewidth]{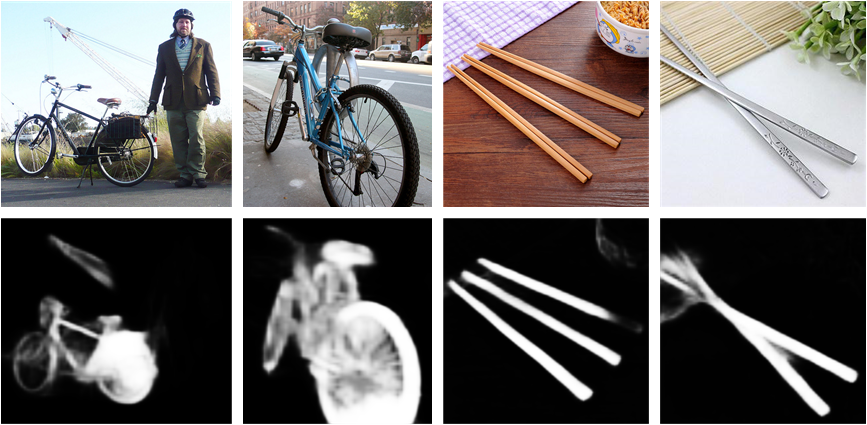}
	\end{overpic}
	\caption{\textbf{Some failure cases} of our model, \eg, ``chopsticks'' and ``bicycles'' with slender or complex structures.
	}
	\label{weakness}
\end{figure}

\begin{figure*}[h]
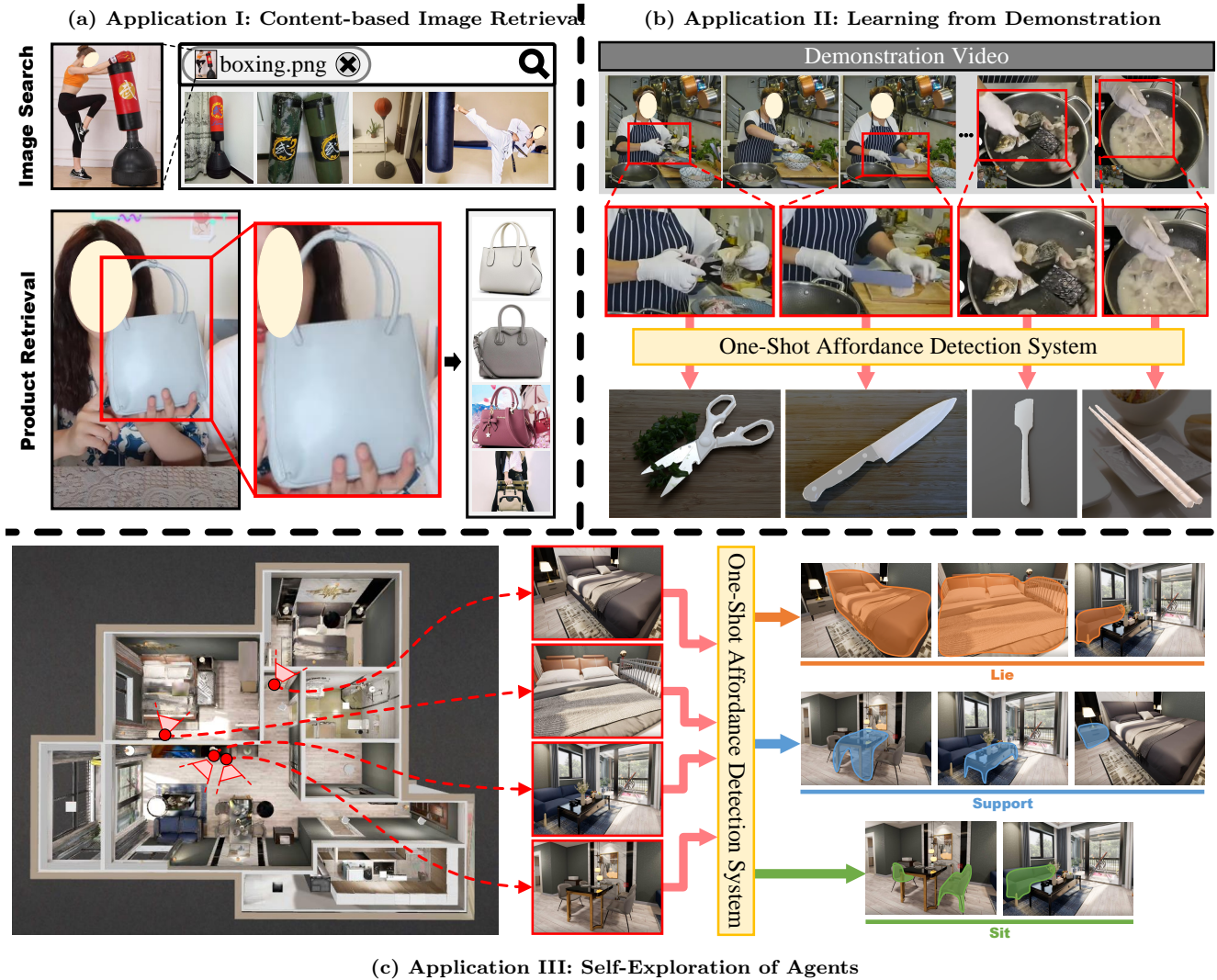

	\centering
		\begin{overpic}[width=1\linewidth]{appall.pdf}
		    \put(5.0, 79.2){\small\textbf{(a) Application \uppercase\expandafter{\romannumeral1}: Content-based Image Retrieval}}
		    \put(52.5, 79.2){\small\textbf{(b) Application \uppercase\expandafter{\romannumeral2}: Learning from Demonstration}}
		    \put(30.0, 0.5){\small\textbf{(c) Application \uppercase\expandafter{\romannumeral3}: Self-Exploration of Agents}}
		\end{overpic}
	\caption{\textbf{Potential application scenarios of the one-shot affordance model.} \textbf{(a) Application \uppercase\expandafter{\romannumeral1}: Content-based Image Retrieval (CBIR).} The CBIR system that has the ability of affordance detection is useful for search engines and online shopping platforms. \textbf{(b) Application \uppercase\expandafter{\romannumeral2}: Learning from Demonstration.} The one-shot affordance detection model can help an agent to select an appropriate tool based on the demonstrator's action. \textbf{(c) Application \uppercase\expandafter{\romannumeral3}: Self-exploration of Agents.} The one-shot affordance detection model can help an agent to automatically recognize the affordance of objects in unknown scenes based on historical data (\eg, images of human interactions).}
	\label{appall}
\end{figure*}

\myPara{Weakness. } Although our model can achieve good results for the one-shot affordance detection task, more efforts should be made in the future work to address the limitations. First, there is still large amount of parameters in the proposed model, which can be further reduced by designing efficient and light-weight backbone networks and modules \citep{GaoEccv20Sal100K,howard2017mobilenets}. Second, it is difficult for the proposed model to segment fine edges or complete object regions when dealing with objects with complex structures (\eg, ``bicycle'') or slender objects (\eg, ``chopsticks''), as shown in Fig. \ref{weakness}. In future work, we can explore a refinement module \citep{Qin_2019_CVPR}, extract high-resolution feature \citep{wang2020deep}, as well as model long-range dependency using transformer block \citep{dosovitskiy2020image} to improve the performance.

\myPara{Potential Applications. }This section discusses several potential applications that could benefit from the research of one-shot affordance detection, including CBIR, learning from demonstration, and self-exploration of agents. Note that we present these applications for illustrative purpose and hope to spark more promising ones in future research.

\begin{itemize}
    
    \item [1)] \textbf{Application \uppercase\expandafter{\romannumeral1}: Content-based Image Retrieval.}
    CBIR \citep{chen2021deep} is an application that searches for semantically matching or appearance similar images in extensive image collections by analyzing their visual content, given a query image that describes the user's intention. As shown in Fig. \ref{appall} (a), the CBIR system that has the ability of affordance detection is useful for search engines (image search) and online shopping platforms (product retrieval) with two advantages. First, it helps to find relevant images with similar affordance. Second, it can help existing image retrieval systems improve retrieval accuracy and diversity by taking the affordance attributes into consideration.

    \item [2)] \textbf{Application \uppercase\expandafter{\romannumeral2}: Learning from Demonstration.}
    Learning from demonstration \citep{argall2009survey}, also known as ``programming by demonstration'', ``learning by imitation'', and ``teaching by demonstration'', has received a great amount of attention in the field of robotics over the last few decades. It aims to drive the robot to learn how to interact with environment by showing the activity of an demonstrator. As shown in Fig. \ref{appall} (b), the one-shot affordance detection model can help an agent to select an appropriate tool based on the demonstrator's action \citep{demo2vec2018cvpr,interaction-hotspots,ho2016generative}. 
    
    \item [3)] \textbf{Application \uppercase\expandafter{\romannumeral3}: Self-exploration of Agents.}
    Self-exploration \citep{ramakrishnan2021exploration} is a key way to empower robots with the perceiving intelligence in unknown scenes. Agents operating in human spaces must master how their environment works: what and how the objects can be used? As shown in Fig. \ref{appall} (c), the one-shot affordance detection model can help an agent automatically recognize the affordance of objects in unknown scenes based on historical data (\eg, images of human interactions \citep{nagarajan2020learning}).
    
\end{itemize}

\myPara{Future Research Directions. }In this part, we discuss several promising research directions for affordance detection.

\begin{itemize}

    \item [1)] \textbf{Multimodal Affordance Detection}
    
    \textbf{$\blacktriangleright$ RGB-D Affordance Detection:} Depth information provides rich spatial structure and layout information, which facilitate extracting geometric clues to improve the performance of affordance detection. Many previous works consider depth information for affordance detection \citep{myers2015affordance,nguyen2017object,zhu2015understanding,sawatzky2017weakly}. Our PADv2 dataset also provides depth information annotations, which can be exploited in future work to estimate human action purpose and capture the intra-class commonality of objects belonging to the same affordance category.
    
    \textbf{$\blacktriangleright$ Vision-Language Affordance Detection:} In this paper, the task of affordance detection is accomplished by mining human action purposes from support images. While in real life, human intentions may also be described by language, \eg, ``want to drink water'', which conveys the information to find objects that can be filled from the scene \citep{chao2015mining,mi2020intention,mi2020intention2}. In the future, it is also promising to consider the marriage of vision and language for the affordance detection task.
    
    \item [2)] \textbf{Affordance Detection from Machine Learning Perspectives}
    
    \textbf{$\blacktriangleright$ Weakly Supervised, Semi-Supervised, and Unsupervised Affordance Detection:} Affordance detection based on supervised learning usually requires large-scale labeled data with pixel-level accurate annotations for training, which are labor-intensive to collect and annotate. Alternatively, weakly supervised, semi-supervised, and unsupervised learning methods are also worth further study for affordance detection
    \citep{sawatzky2017adaptive,interaction-hotspots,pan2021unveiling,wang2021minmaxcam,gao2021ts}.
    
    \textbf{$\blacktriangleright$ Affordance Detection from Imbalanced Data:} Most of the existing affordance datasets have imbalanced data distribution from real-world scenes, \eg, some long-tailed categories only have a small number of samples. Data imbalance poses a significant challenge to the task of affordance detection, since the model has to learn a fast adapting ability as well as overcoming the forgetting problem due to the insufficient amount of training data of long-tailed affordance categories. Nevertheless, it is of great practical significance and worth more research efforts \citep{chen2019residual,wu2019iou,qian2020dr,he2009learning}.
    
    \item [3)] \textbf{Real-world Applications and Systems}
    
    Some potential applications (CBIR \citep{chen2021deep}, learning from demonstration \citep{demo2vec2018cvpr,interaction-hotspots}, and self-exploration of agents \citep{nagarajan2020learning}) have already been discussed in the previous section. It is interesting to explore how practical systems can be enhanced by incorporating the affordance detection model. Moreover, developing an embodied agent that can continuously learn to perceive object afforance in real-world environment and interact with them to accomplish specific tasks could also be a promising research topic.
    
\end{itemize}

\newcommand{\tabincell}[2]{\begin{tabular}{@{}#1@{}}#2\end{tabular}}  


\normalem
\bibliographystyle{spbasic}      
{\bibliography{IJCV}}

\end{document}